\documentclass{article}

\usepackage{microtype}
\usepackage{graphicx}

\usepackage{enumitem}
\usepackage{footnote}
\makesavenoteenv{tabular}
\makesavenoteenv{table}
\usepackage{tikz}
\tikzset{font={\fontsize{8pt}{12}\selectfont}}
\usetikzlibrary{decorations.pathreplacing}
\usetikzlibrary{shapes.arrows}
\usetikzlibrary{shapes,arrows, matrix}

\usepackage[caption=false]{subfig}
\usepackage{booktabs} 
\usepackage{amsfonts}
\usepackage{amsmath}
\usepackage{amsthm}
\usepackage{mathabx}

\usepackage{multirow}

\usepackage{graphicx}
\usepackage{enumitem}
\setlist{nosep}

\usepackage{hyperref}
\usepackage{dsfont}

\newcommand{\BBr}{\mathbb{R}}

\usepackage[accepted]{icml2024}

\icmltitlerunning{
Combining Statistical Depth and Fermat Distance for Uncertainty Quantification
}

\numberwithin{equation}{section}
\numberwithin{table}{section}
\numberwithin{figure}{section}

\begin{document}

\setlength{\abovedisplayskip}{1pt}
\setlength{\belowdisplayskip}{1pt}

\twocolumn[
\icmltitle{
Combining Statistical Depth and Fermat Distance for Uncertainty Quantification
}



\icmlsetsymbol{equal}{*}

\begin{icmlauthorlist}
\icmlauthor{Hai-Vy Nguyen}{rn,imt,irit}
\icmlauthor{Fabrice Gamboa}{imt}
\icmlauthor{Reda Chhaibi}{imt}
\icmlauthor{Sixin Zhang}{irit}
\icmlauthor{Serge Gratton}{irit}
\icmlauthor{Thierry Giaccone}{rn}
\end{icmlauthorlist}

\icmlaffiliation{rn}{Ampere Software Technology}
\icmlaffiliation{imt}{Institut de mathématiques de Toulouse}
\icmlaffiliation{irit}{Institut de Recherche en Informatique de Toulouse}

\icmlcorrespondingauthor{Hai-Vy Nguyen}{hai-vy.nguyen@renault.com}

\icmlkeywords{out-of-distribution, uncertainty, ICML}

\vskip 0.3in
]



\printAffiliationsAndNotice{}  

\begin{abstract}
We measure the Out-of-domain uncertainty in the prediction of Neural Networks using a statistical notion called ``Lens Depth'' (LD) combined with Fermat Distance, which is able to capture precisely the ``depth'' of a point with respect to a distribution in feature space, without any assumption about the form of distribution. Our method has no trainable parameter. The method is applicable to any classification model as it is applied directly in feature space at test time and does {\it not} intervene in training process. As such, it does {\it not} impact the performance of the original model. The proposed method gives excellent qualitative result on toy datasets and can give competitive or better uncertainty estimation on standard deep learning datasets compared to strong baseline methods.
\end{abstract}

\setlength{\abovedisplayskip}{3pt}
\setlength{\belowdisplayskip}{1pt}

\newtheorem{prop}{Proposition}

\section{Introduction}\label{intro}

We consider a multi-class classification problem with the input space $\mathcal{X}$. In general, a classification model consists of a feature extractor (backbone) $\Phi_{\theta_1}$ and a classifier $h_{\theta_2}$: $f_{\theta} = h_{\theta_2} \circ \Phi_{\theta_1}$, where $\theta = \theta_1 \cup \theta_2$ is the set of parameters of the model. The backbone transforms inputs into fixed-dimension vectors in the so-called \textit{feature space} $\mathcal{F}$.
The classifier $h$ then maps the features to predictions. In our experiments, the classification model is provided by a neural network, $h$ is a softmax layer consisting of a linear transformation and a \textit{softmax} function, $\mathcal{F}$ is the output space of the penultimate layer right before the softmax layer. The model $f_{\theta}$ is trained on  i.i.d.  examples drawn  from \textit{In-Distribution} (ID) $P_{\text{in}}$.  $f_{\hat{\theta}}$ denotes the trained model.

\begin{figure}
  \centering
  \subfloat[Gaussian Prior]{\label{fig:two_moon_gaussian_intro}\includegraphics[width=0.3\columnwidth]{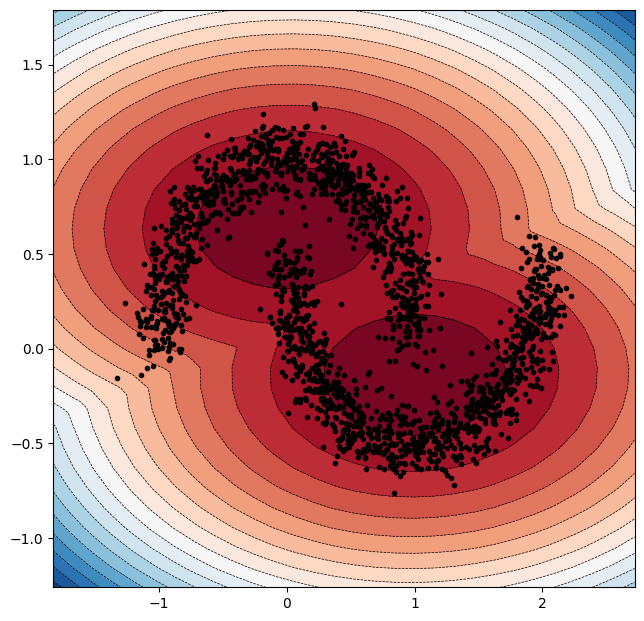}}
  \subfloat[Our method]{\label{fig:two_moon_LD}\includegraphics[width=0.3\columnwidth]{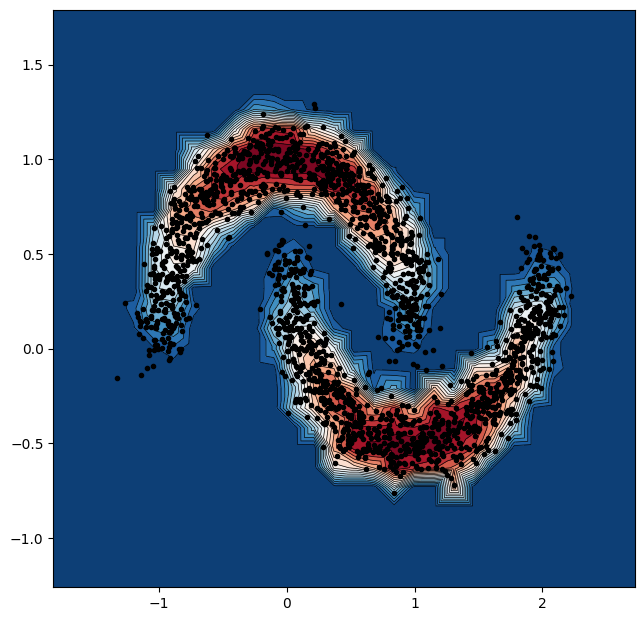}}
\caption{
Motivation example using the two-moons dataset. Colors indicate OOD score. Gaussian fitting (left), fails completely to capture the distribution of dataset whereas our proposed method (right) represents very well how central a point is  with respect to (w.r.t.) clusters without any \textit{prior assumption}.
} 
\label{fig:two_moon_intro}
\vspace{-1\baselineskip}
\end{figure}

\textbf{OOD detection.} Classification neural networks have proved highly effective in terms of precision. However, beside performance, in critical applications, one needs to detect out-of-distribution (OOD) data for safety reasons. Indeed, at the inference stage, the model should only predict for data coming from the ID and reject OOD samples. For this purpose, one needs to associate a confidence (or uncertainty) score $S$ with these data so that one can reject uncertain predictions. This is  referred as \textit{Out-of-domain uncertainty} \cite{gawlikowski2023survey}. At the inference stage, $x$ is considered as ID if $S(x)\geq \varepsilon$ (with some threshold $\varepsilon \in \BBr$) and OOD otherwise.



We develop a method applicable directly in the feature space $\mathcal{F}$ of the trained model $f_{\hat{\theta}}$. It yields a score function $S_{\mathcal{F}}$: $S(x):= S_{\mathcal{F}}(\Phi_{\hat{\theta}_1}(x))$. One major advantage is that there is no need of supplementary training. Besides, the model's performance is also preserved. Discussion about advantages of this approach over other methods is in Section \ref{related}.

A desirable property for any well-trained model is the preservation of the data geometry in the feature space.In other words, in this space, similar data should be close and dissimilar data should be far away. Intuitively, each class should be represented as a different cluster in feature space (see, e.g. \citet{caron2018deep}, \citet{yang2017towards}).
In addition, data that are dissimilar to the training data should be distant from any cluster obtained on the training data in the feature space. Assuming this desirable property, a method  measuring directly ``how central'' a point is with respect to (w.r.t.) clusters taking into account density and geometry of each cluster in the feature space  should provide an uncertainty score. For this objective, standard methods consist in assuming some \textit{prior} distribution such as a Gaussian model \cite{lee2018simple}.  However, the assumption that the data in a cluster is Gaussian distributed or follow any particular distribution is quite restrictive. We will show in our experiments section that the Gaussian assumption fails even in a very simple case (Section \ref{sec:two_moon}). Let us take the example of a  simple frame in the plane  with 2 clusters corresponding to 2 classes in form of two-moons (Fig. \ref{fig:two_moon_intro}).
In this example, Gaussian fitting (Fig. \ref{fig:two_moon_gaussian_intro}) fails totally to capture the distribution of clusters.

\begin{figure}[H]
\centering
\begin{tikzpicture}[scale=1.0, 
    every node/.style={fill=white}, align=center]


\node[draw, rectangle, minimum width=2.5cm, 
                        minimum height=2.5cm,
                        dashed, rounded corners=0.5cm] (nn) at (0,-1.6) {};
\node[right of=nn, xshift=2cm, yshift=2cm] (nn_text) {Model trained with in-distribution (ID) data};
\node[draw, ellipse, minimum width=10mm, minimum height=5mm, align=left, text width=10]  (input) at (0, 0) {$X$};
\node[draw, trapezium, minimum width=10mm, minimum height=10mm, align=center, trapezium left angle=120, trapezium right angle=120, align=center, below of=input, yshift=-0.2cm]  (feature) {$\Phi$};
\node[draw, rectangle, fill=lightgray, align=center, minimum width=10mm, below of=feature, yshift=0.4cm] (feature_layer) {};
\node[draw, rectangle, minimum width=10mm, minimum height=7mm, below of=feature_layer, yshift=0.55cm]  (classifier) {$h$};
\node[draw, ellipse, below of=classifier, yshift=-0.05cm]  (classification_task) {Classification};

\node[xshift=0.8cm, right of=feature_layer, xshift=0.0cm, minimum height=1cm] (feature_vector) {$\Phi(X) \in \mathcal{F}$ \\ (Features)};
\node[draw, ellipse, minimum height=1.6cm, 
                     minimum width =1.8cm, opacity=0.0, draw opacity=1.0] at (feature_vector) {  };

\node[draw, right of=feature_vector, rounded corners=.3cm, xshift=1.0cm, yshift=0.9cm]  (ld) {Lens depth \\ as Eq. \eqref{eq:emp_ld}};
\node[draw, below of=ld, rounded corners=.3cm, yshift=-0.9cm]  (fermat) {Fermat distance \\ as Eq. \eqref{eq:modified_sample_fermat}};
\node[draw, ellipse, minimum height=1.6cm, 
                     minimum width =1.9cm, 
                     right of=feature_vector, xshift=3.0cm, yshift=0cm]  (score) {};
\node[] at (score) {Score \\ as Eq. \eqref{eq:define_score}};

\tikzset{south_arrow/.style={single arrow, draw, fill=lightgray, minimum width=10mm, minimum height=10mm, scale=0.5,
                         inner sep=0ex, single arrow head extend=0.1mm, rotate=-90}
}

\tikzset{east_arrow/.style={single arrow, draw, fill=lightgray, minimum width=10mm, minimum height=20mm, scale=0.5,
                         inner sep=0ex, single arrow head extend=0.1mm}
}

\tikzset{east_arrow_long/.style={single arrow, draw, fill=lightgray, minimum width=50mm, minimum height=100mm, scale=0.5,
                         inner sep=0ex, single arrow head extend=0.1mm}
}

\path (input)         ++(-0.07,-0.45) node[anchor=south, south_arrow, scale=0.8] (feature) {};
\path (classifier)    ++(-0.07,-0.5) node[anchor=south, south_arrow, scale=0.8] (classification_task) {};

\path (feature_layer) ++(0.9,0) node[anchor=east, east_arrow, scale=0.4] (feature_vector) {};

\draw (feature_vector.east) ++(2.7,0) node[east_arrow_long, scale=0.45] (score.west) {};

\draw[->] (nn_text) ++(0,-0.2) -- (nn) {};

\end{tikzpicture}
\caption{General scheme of our method. 
Given a set of features $\Phi$, the Fermat distance is a metric which respects and adapts to the distribution of $\Phi$. Lens depth wraps the Fermat distance into a probabilistic and interpretable score $S$.
No additional training is needed.
} 
\label{fig:general_scheme_tikz}
\vspace{-1.\baselineskip}
\end{figure}
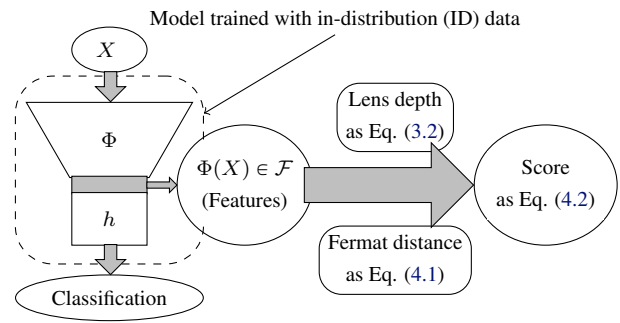

This motivates us to develop a non-parametric method that can measure explicitly how ``central'' a point is w.r.t. a cluster without the need of additional training and prior assumption.  Furthermore, the method should accurately capture distribution with complex support and shape, in order to be adapted to a variety of cases. To measure how central a point is w.r.t. a distribution, we use the so-called notion of statistical  \textit{Lens Depth} (\textit{LD}) \cite{liu1990notion}, that  will be presented in Section \ref{sec_LD}. Furthermore, for \textit{LD} to correctly capture the shape of the distribution, an appropriate distance must be adopted that adaptively takes into account its geometry and density. Fermat distance is a good candidate for this purpose. However, it is not directly tractable as it stands on integrals along rectifiable paths. A recent paper \citet{groisman2022nonhomogeneous} proposes the use of  an explicit sampled Fermat distance and shows its consistency property (see also \citet{cholaquidis2023weighted}). In our work, we make use of their results to compute the \textit{LD}. The  general scheme is illustrated in Figure \ref{fig:general_scheme_tikz}.

\textbf{Consistency of the uncertainty score.} \textit{Out-of-domain} uncertainty is more demanding than OOD detection. A consistent uncertainty score function should allow us to detect OOD.  Furthermore, when more samples are rejected based on this score, the accuracy of the  multi-class classification on the retained samples should increase. In other words, the fewer examples retained (based on the score), the better the accuracy. This means that the score represents the true confidence in prediction. Many previous methods (e.g. \citet{sun2022out}, \citet{liang2017enhancing}) focus purely on OOD detection, and this second desired property is not satisfied or proved. In contrary, we show in Sections \ref{sec:fashion_mnist} and \ref{sec:svhn_cifar10} that our method respects this property. This is because our method measures a natural "depth" of the considered example. Consequently, the larger the depth of this example, the more typical this point is (relative to the training set), and so the easier it is for the model to classify. 

In summary, our contribution is at the following three levels:
\begin{itemize}[noitemsep,topsep=-4pt]
    \item We are bringing to machine learning the mathematical tool of $LD$, combined with Fermat distance.
    It proves particularly efficient for OOD uncertainty quantification.
    We also propose improvements that avoid undesirable artifacts, and simple strategies for reducing significantly the complexity in computing $LD$. 
    \item The method we propose is non-parametric and non-intrusive. We do not have priors on the distribution of data nor features. We do not require modifying the training algorithms.
    \item The method is almost hyperparameter-free, as we show that it is rather insensitive the only parameter used to define Fermat distance.
\end{itemize}

Tables \ref{table:auroc_mnist} and \ref{table:auroc_cifar10} in the experiments section give benchmarks. And 
our code can be shared upon request. 

%
%

\section{Related Work}
\label{related}

One approach to construct a confidence score consists in fine-tuning the model $f_{\hat{\theta}}$ using some auxiliary OOD data so that the ID and OOD data are more separable in the feature space \cite{liu2020energy}.  One may  even use very particular type of model and training mechanism for the original classification task such as RBF models \cite{lecun1998gradient} in DUQ method \cite{van2020uncertainty} or \textit{Prior Networks} in which the prior distribution is put on on the output of the model \cite{malinin2018predictive}. More laborious method to handle uncertainty in neural network is its Bayesian modeling ( \citet{MacKay1992APB}, \citet{gal2016uncertainty}). Another approach is to train some additional models to train such Deep Ensembles \cite{lakshminarayanan2017simple} or LL ratio \cite{ren2019likelihood}. In these approaches, one needs to carefully perform the supplementary training. Otherwise one could reflect wrongly the true underlying distribution. Moreover,  the performance of the multi-class classification task could be impacted. Furthermore, the training process itself contains some stochastic factors (e.g. mini-batch sampling), that could increase the error of the confidence score. For all these reasons, these methods can be considered \textit{intrusive}.



Independently from above methods, a \textit{non-intrusive} approach is to work directly in the feature space $\mathcal{F}$ of the trained model $f_{\hat{\theta}}$. This is \textit{non-intrusive} in the sense that there is no need of supplementary training. Besides, model performance is not impacted.
One of the simplest method is to use the $k$-nearest neighbor distance \cite{sun2022out}. It is very simple but has the drawback of completely ignoring  cluster geometry and density.  A more sophisticated approach uses minimum Mahalanobis distance \footnote{The Mahalanobis distance from a point $x \in \BBr^d$ to a given probability $Q$ (with mean $\mu$ and covariance matrix $\Sigma$) is defined as $d(x,Q)= \sqrt{(x-\mu)^T \Sigma^{-1}(x-\mu)}$.} \cite{mahalanobis2018generalized} based on \textit{Gaussian prior} \cite{lee2018simple}. Despite taking the distribution into account, Gaussian modeling is too restrictive as it leads to an ellipsoid for shaping each cluster. A classical non-parametric method is one-class SVM \cite{scholkopf1999support} which demands supplementary training. Besides, its effectiveness is known to be significantly influenced by  the kernel choice and its hyper-parameters. On the other hand, our method needs no additional training and has following properties at the same time: (1) \textit{non-intrusive}; (2) non-parametric; (3) nearly hyperparameter-free. All these properties make it unique in comparison to other methods.

\section{Background}
\label{background}
\subsection{Lens Depth} \label{sec_LD}
Lens depth (LD) \cite{liu1990notion} is a specific notion of a more general quantity called Depth \cite{tukey1975mathematics}. A depth is a score measure  of the membership  of a point w.r.t. a distribution in a general space. The greater the depth, the more central the point is to the distribution. LD of a point w.r.t a distribution $P_X$ is defined as the probability that it belongs to the intersection of two random balls. These balls are centered at two independent random points $X$ and $Y$, both having the distribution $P_X$ and a radius equal to the distance between $X$ and $Y$. More formally, if we work on $\BBr^d$, the LD of a point $x \in \BBr^d$ w.r.t. $P_X$ is defined as follows,
\begin{equation}
\begin{aligned}
 LD(x,P_X) := \mathbb{P}(x \in B_1 \cap B_2) \ .
\end{aligned}
\label{eq:theory_LD}
\end{equation}
Here 
\begin{itemize}
\setlength\itemsep{-0.1em}
    \item $d$ is a given distance on $\BBr^d$,
    \item $X_1$,$X_2$  are i.i.d  with law $P_X$,
    \item $B(p, r)$ is the closed ball centered at $p$ with radius $r$,
    \item $B_i=B(X_i,d(X_2,X_1)),\ i=1,2$.
\end{itemize}

Let  $A(X_1,X_2) = B_1 \cap B_2$.
Equation \eqref{eq:theory_LD} naturally gives rise to the following empirical version of LD, 
\begin{equation}
\widehat{LD}_n(x) := {\binom{n}{2}}^{-1} \sum_{1\leq i_1 < i_2 \leq n} \mathds{1}_{A(X_{i_1},X_{i_2})}(x) \ .
\label{eq:emp_ld}
\end{equation}
Note that for the empirical version, the intersection set might be rewritten as
\begin{equation}
    A(X_1,X_2) = \{ x 
                    : \max_{i=1,2} d(x,X_i)\leq d(X_1,X_2)\} \ .
\end{equation}

\begin{figure}[ht]
\centering
  \subfloat[Gaussian]{\label{fig:ld_naive_gaussian}\includegraphics[width=0.25\columnwidth]{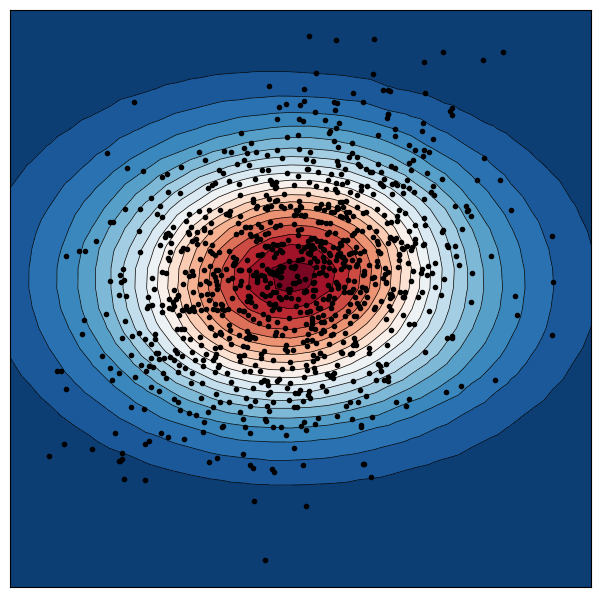}}
  \subfloat[Moon]{\label{fig:ld_naive_moon}\includegraphics[width=0.25\columnwidth]{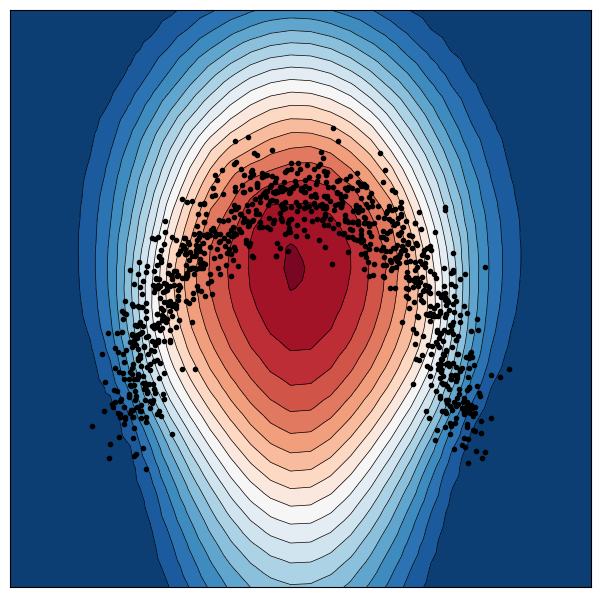}}
\caption{$\widehat{LD}$ using Euclidean distance. We see that using simply Euclidean distance cannot capture correctly the distribution.}\label{fig:ld_naive}
\end{figure}

Obviously, a crucial question is the choice of the distance $d$. A naive choice is the Euclidean one. Examples of $\widehat{LD}$ using Euclidean distance are depicted in Fig. \ref{fig:ld_naive}. We see that in the Gaussian case, the level curves of $\widehat{LD}$ rather well capture the distribution. However, for the moon distribution they fail miserably. This is not surprising as the Euclidean distance does not take into account the data distribution $P_X$. 

This gives rise to a natural problem as stated by \citet{groisman2022nonhomogeneous}: \textit{How to learn a distance that can capture both the geometry of the manifold and the underlying density?} The Fermat distance allows us to solve this problem and it is presented in the following section.

\subsection{Fermat distance}

Following \citet{groisman2022nonhomogeneous}, let $S$ be a subset of $\mathbb{R}^d$. For a continuous and positive function $f: S \rightarrow \mathbb{R}_+$ , $\beta \geq 0$ and $x,y \in S$, the Fermat distance $\mathcal{D}_{f,\beta}(x,y)$ is defined as 
\begin{equation}
\mathcal{D}_{f,\beta}(x,y) := {\inf}_{\gamma}\mathcal{T}_{f,\beta}(\gamma) \ ,
\end{equation}
where
\begin{equation}
\mathcal{T}_{f,\beta}(\gamma) := \int_\gamma f^{-\beta} \ .
\end{equation}

The infimum is taken over all continuous and rectifiable paths $\gamma$ contained in $\bar{S}$, the closure of $S$, that start at $x$ and end at $y$.

\textbf{Sample Fermat Distance.} Let $Q$ be a non-empty, locally finite, subset of $\mathbb{R}^d$, serving as dataset. $|x|$ denotes Euclidean norm of $x$, $q_Q(x) \in Q$ is the particle closest to $x$  in Euclidean distance -- assuming uniqueness\footnote{Of course, uniqueness is generically achieved, for example almost surely for random points sampled according to a diffuse measure.}. 
For $\alpha \geq 1$, and $x,y \in \mathbb{R}^d$, the sample Fermat distance is defined as:
\begin{align}
  & D_{Q,\alpha}(x,y) := \min \Big\{
  \sum_{j=1}^{k-1} |q_{j+1}-q_j|^\alpha \ :   (q_1, \dots,  q_k)\in Q^k 
  \nonumber \\ 
  & \quad \quad \quad \text{ with } q_1=q_Q(x), \ q_k = q_Q(y), \ k \geq 1
  \Big\} \ .
\label{eq:sample_fermat}
\end{align}

\citet{groisman2022nonhomogeneous} show that the sample Fermat distance when appropriately scaled converges to the Fermat distance.

\begin{figure}
  \centering
  \subfloat[$\alpha=1$]{\label{fig:min_path_1}\includegraphics[width=0.25\columnwidth]{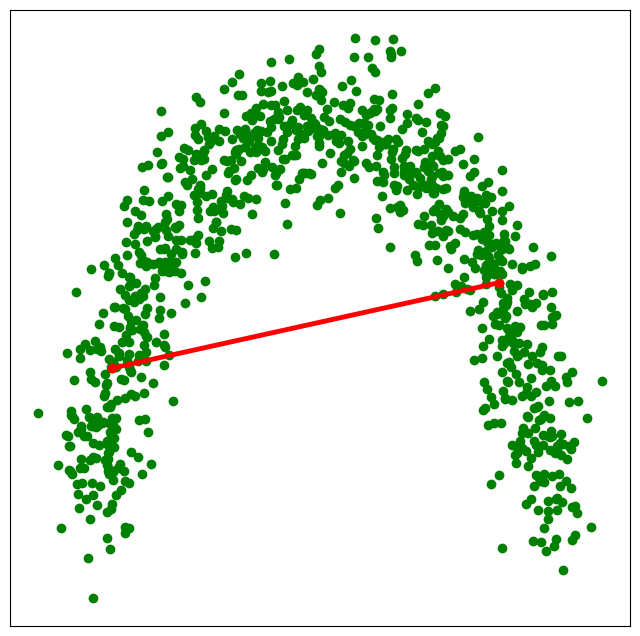}}
  \subfloat[$\alpha=1.2$]{\label{fig:min_path_1.2}\includegraphics[width=0.25\columnwidth]{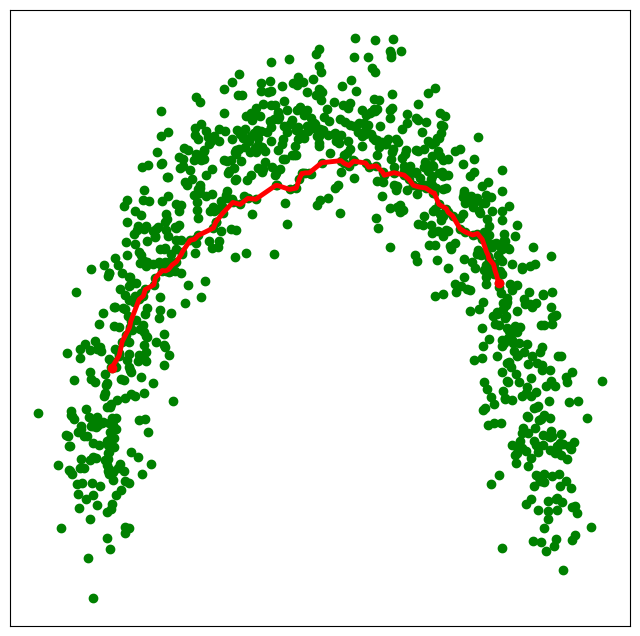}}
  \subfloat[$\alpha=3$]{\label{fig:min_path_3}\includegraphics[width=0.25\columnwidth]{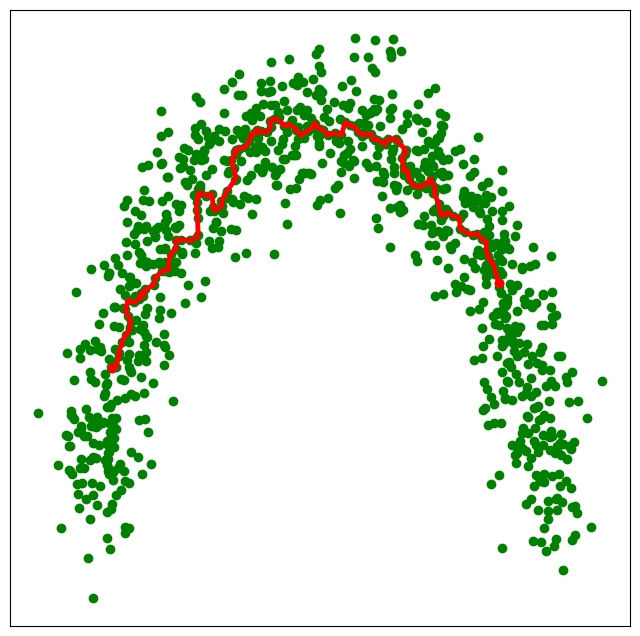}}
  \subfloat[$\alpha=7$]{\label{fig:min_path_7}\includegraphics[width=0.25\columnwidth]{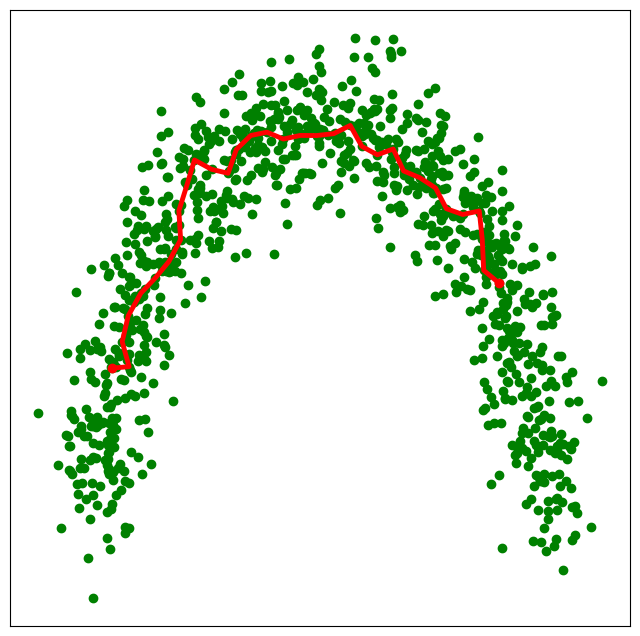}}
\caption{Sample Fermat path between two fixed randomly chosen points using different values of $\alpha$.} 
\label{fig:min_path}
\vspace{-1.\baselineskip}
\end{figure}

\textbf{Intuition behind Sample Fermat Distance.} The sample Fermat distance searches for the shortest path relating the points. The length of each path is the sum of the Euclidean distances of consecutive points in the path powered by $\alpha$. With $\alpha=1$, the shorted path between $x$ and $y$ is simply  the line relating $q_Q(x)$ and $q_Q(y)$ (Fig. \ref{fig:min_path_1}). However, with a sufficiently large $\alpha$, this definition of path length discards consecutive points with a large Euclidean distance instead favoring points that are closely positioned in terms of Euclidean distance. So, this will qualify the path passing through high density areas.  Moreover, as this distance depends also on the number of terms in the sum in Eq. \eqref{eq:sample_fermat}, this enforces a path to be smooth enough. These two remarks show that Fermat distance naturally captures the density and geometry of the data set.

In Fig. \ref{fig:min_path}, we go back to the moon example. We randomly choose  2  points and compute the Fermat path. We see that with $\alpha=1$, we recover the Euclidean distance and so the Fermat path is simply a line. For $\alpha$ larger than 1 but not large enough (for instance, $\alpha=1.2$, Fig. \ref{fig:min_path_1.2}), the Fermat path still does not capture the orientation of the dataset. However, as $\alpha$ gets larger, the Fermat path rapidly tracks the orientation of the dataset. For instance, with $\alpha=3$, the path follows  very well the distribution shape.

\section{Combining LD and Fermat Distance}
\label{FLD}

\subsection{Artifacts from classical Fermat distance}
In the computation of the depth, instead of using Euclidean distance, we use sample Fermat distance. The results for the moon and spiral datasets are depicted in Fig. \ref{fig:ld_fermat_naive} and \ref{fig:ld_fermat_naive_spiral}. We see that the shape of datasets is much better captured. However, we also observe some zones having constant LD  value (represented by the same color). The existence of such zones are explained by the following proposition:

\begin{figure}
\centering
\subfloat[Before \label{fig:ld_fermat_naive}]{\includegraphics[width=0.25\columnwidth]{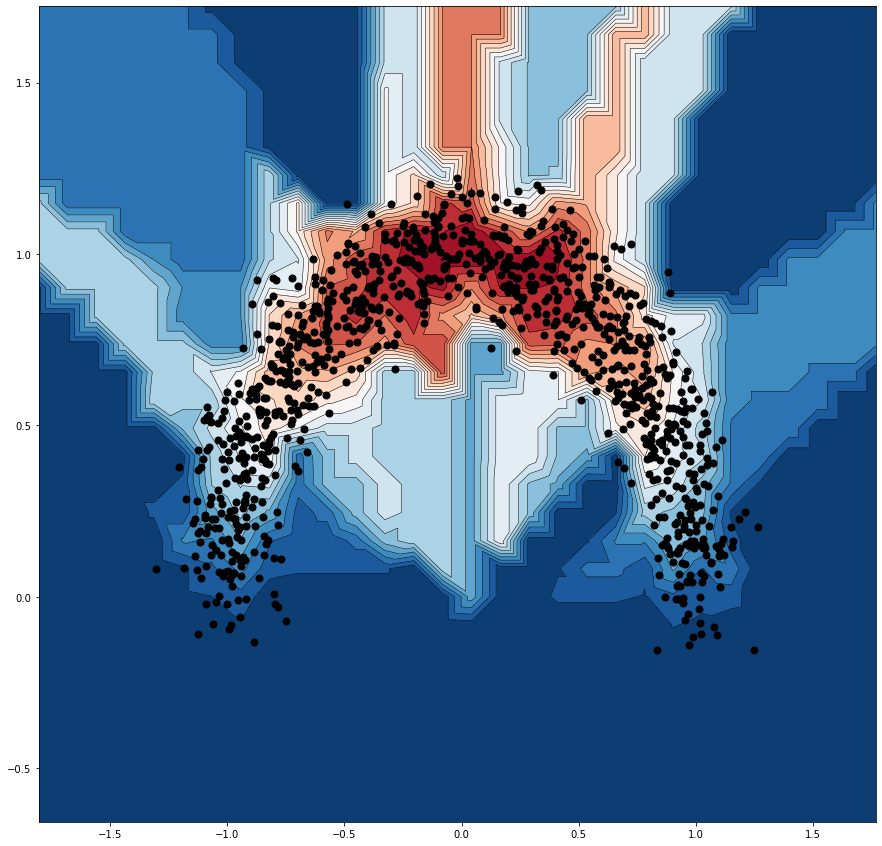}}
\subfloat[Before \label{fig:ld_fermat_naive_spiral}]{\includegraphics[width=0.25\columnwidth]{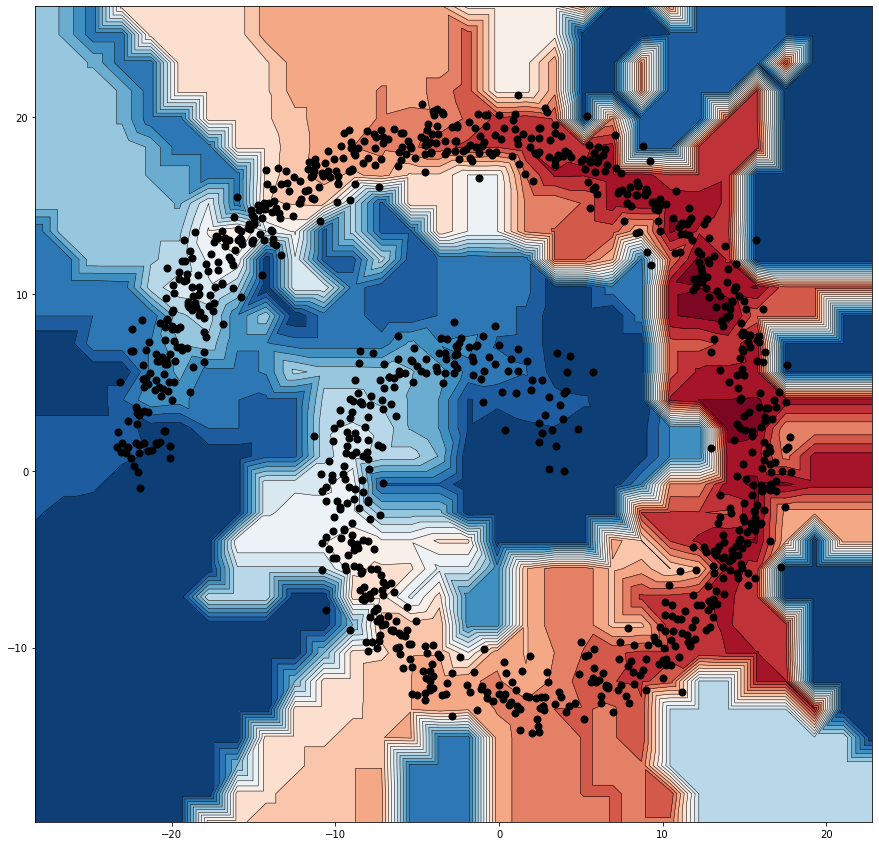}}
\subfloat[After \label{fig:ld_fermat_moon}]{\includegraphics[width=0.25\columnwidth]{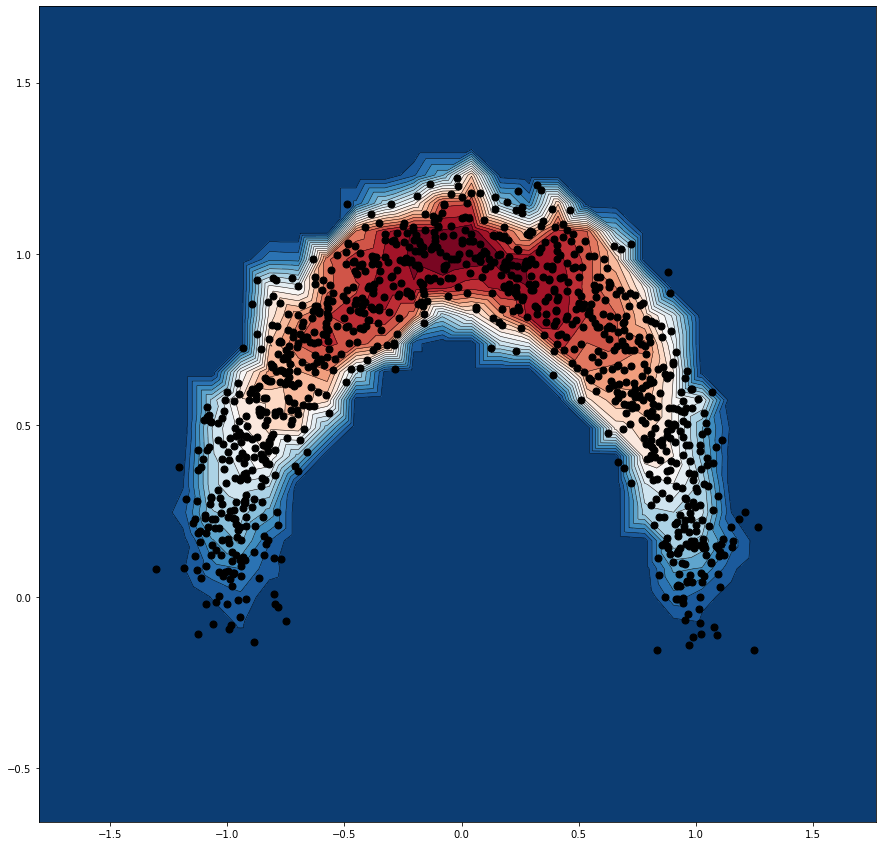}}
\subfloat[After \label{fig:ld_fermat_spiral}]{\includegraphics[width=0.25\columnwidth]{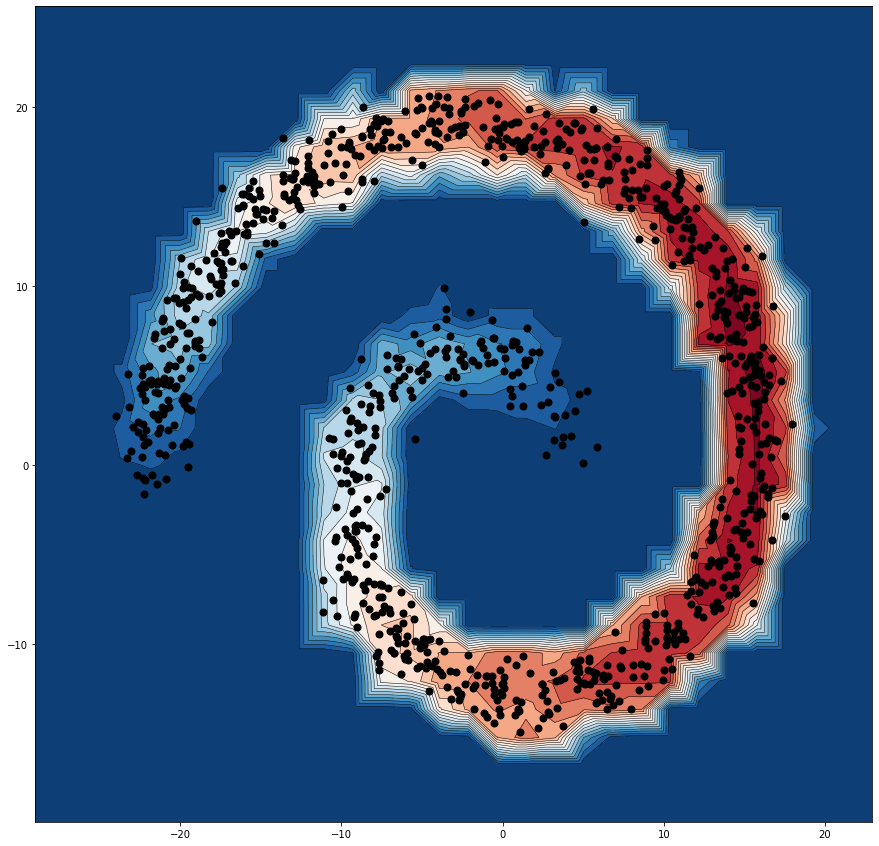}}
\caption{$\widehat{LD}$ with Sample Fermat Distance on moon and spiral datasets. (a) and (b) are results of using directly sample Fermat distance in Eq. \eqref{eq:sample_fermat}. This produces undesirable artifacts where we observe zones of constant value of $LD$. This phenomenon is explained by Proposition \ref{prop:1}. (c) and (d) use our modified version in Eq. \eqref{eq:modified_sample_fermat}: it captures perfectly the distributions.} \label{fig:ld_fermat}
\vspace{-0.5\baselineskip}
\end{figure}

\begin{prop}
For $x\in\BBr^d, \;\widehat{LD}(x)=\widehat{LD}(q_Q(x))$. 
In other words, the empirical lens depth is constant over the Vorono\"i cells\footnote{Definition of the Vorono\"i cells is in \textbf{Appendix} \ref{voronoi}.} associated to $Q$.  
\label{prop:1}
\end{prop}

The proof of Proposition \ref{prop:1} is in \textbf{Appendix} \ref{proof}.
The consequence of the last proposition is that, even for a point far removed  from $Q$, the value of $\widehat{LD}$ remains the same as that of its nearest point in $Q$. Consequently, $\widehat{LD}$ does not vanish at infinity. This is totally undesirable, as an ideal property of any depth is to vanish at infinity.
To avoid this undesirable artifact, we need to modify the sample Fermat distance so that it takes into account the distance to $Q$.

\subsection{Modified Sample Fermat Distance}
The modified distance is defined, for $y\in Q,\; x\in\BBr^d$ as follows:
\begin{equation}
D^{\mathrm{modif}}_{Q,\alpha}(x,y) := \min_{q\in Q}\{|x-q|^{\alpha} + D_{Q,\alpha}(q,y)\} \ . 
\label{eq:modified_sample_fermat}
\end{equation}

Here, $D_{Q,\alpha}(q,y)$ has been defined in Eq. \eqref{eq:sample_fermat}.

\textbf{Interpretation.} In the original definition in Eq. \eqref{eq:sample_fermat}, the path always starts by the closest point in the dataset. Consequently, the distance to this closest point is totally ignored. 
To eliminate this drawback, the distance to a potential starting point lying in $Q$ is added. Note that  the optimization problem for calculating $D^{\mathrm{modif}}_{Q,\alpha}$ is of the same type as that for calculating $D_{Q,\alpha}$ with only a change of starting point. Hence, the consistency of this empirical distance towards the theoretical Fermat distance remains true. Indeed,  in the new  formulation \eqref{eq:modified_sample_fermat},  the point $q\in Q$ is not fixed at $q_{Q}(x)$ but remains free and is a part of the optimization problem. Notice further that  our modified version enjoys two nice properties. Firstly,  if $x\in Q$  then Eq. \eqref{eq:modified_sample_fermat} coincides with Eq. \eqref{eq:sample_fermat} ($D^{\mathrm{modif}}_{Q,\alpha}(x,y)= D_{Q,\alpha}(x,y)$). Secondly, $D^{\mathrm{modif}}_{Q,\alpha}(x,y)$ increases to infinity when $x$ is going far away from $Q$. Consequently, in this case, the corresponding $\widehat{LD}$ w.r.t $Q$  tends to $0$. 
The $\widehat{LD}$ using this modified version of the distance is displayed  on two examples in Fig. \ref{fig:ld_fermat_moon} and \ref{fig:ld_fermat_spiral}. 
With our modification, the undesirable artifact of constant-valued zone is erased. Furthermore,  for points far away from the dataset, $\widehat{LD}$ tends quickly to 0. In conclusion, our method captures the shape of distributions perfectly.


\subsection{Qualitative evaluation of stability}
We experiment and evaluate the stability of our method on 
the spiral dataset. This is a tricky dataset, and a standard method like the Gaussian one  cannot capture its shape.

\subsubsection{Stability with respect to number of training points}
When running a statistical algorithm, it is desirable to have as large a sample as possible. 
However, in many cases, only a very small amount of data is available. This motivates the study of the stability of our method in a small data regime. Here, we simulate the spiral dataset with 1000 points. Then, we choose randomly only $20\%$ of the simulated points (i.e. 200 points) as the sample dataset to compute $LD$. We perform different runs for different random samples with $\alpha=5$ for a visual evaluation. For the sake of brevity, only the results of four tests are shown in Fig. \ref{fig:stab_points}. More replications are displayed in \textbf{Appendix} \ref{sec:append_stable_points}. We see that in the 4 tries, our method gives slightly different estimation of $LD$. This small fluctuation is to be expected, as we take only $20\%$ of the points at random each time.  Nevertheless, the method captures the shape of the data set really well (the full sample of 1000 points is displayed in the figures). 

\begin{figure}[ht]
  \centering
  \subfloat[$1^{st}$ try]{\label{fig:stab_points_1}\includegraphics[width=0.25\columnwidth]{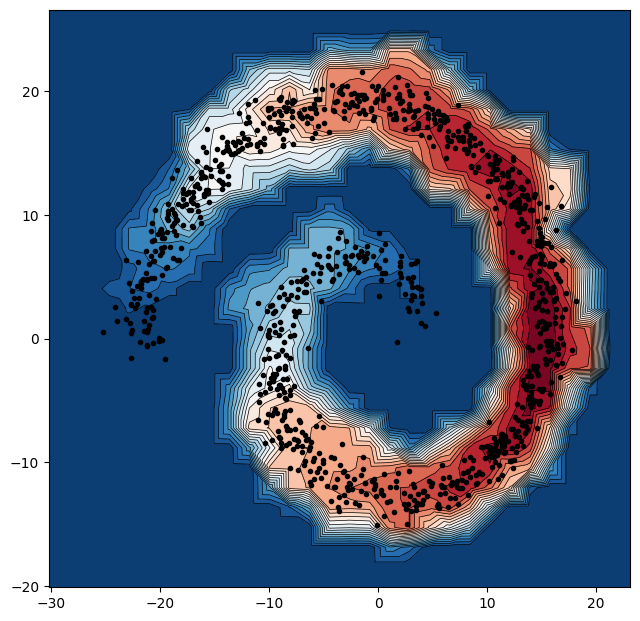}}
  \subfloat[$2^{nd}$ try]{\label{fig:stab_points_2}\includegraphics[width=0.25\columnwidth]{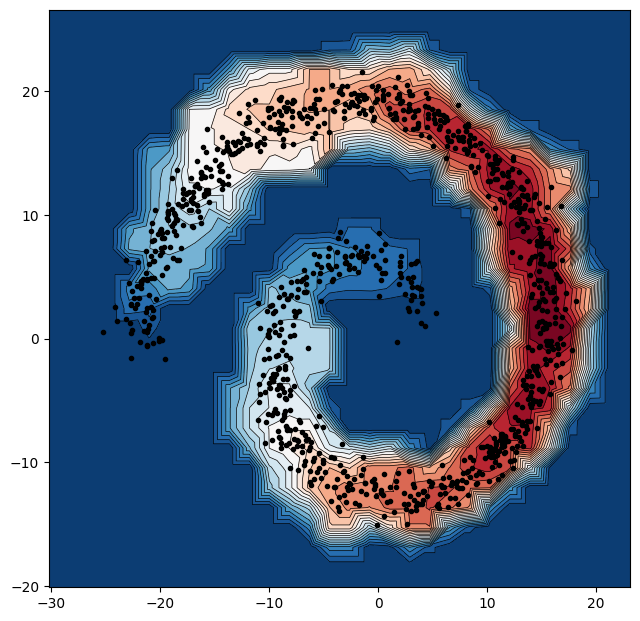}}
  \subfloat[$3^{rd}$ try]{\label{fig:stab_points_3}\includegraphics[width=0.25\columnwidth]{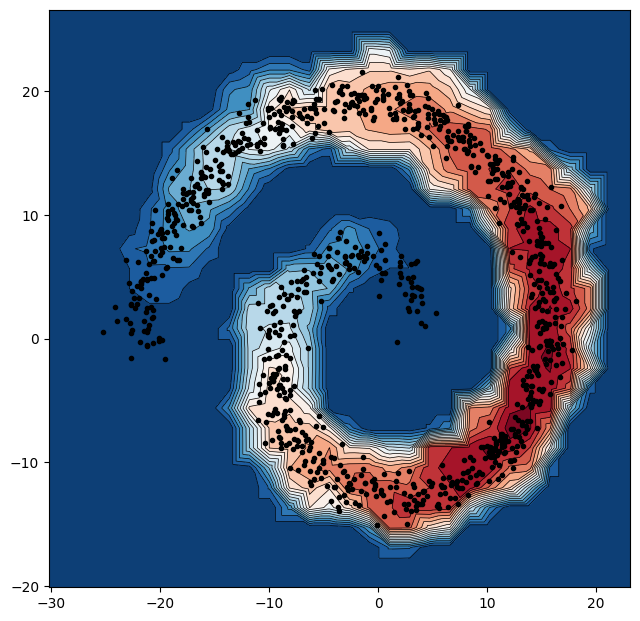}}
  \subfloat[$4^{th}$ try]{\label{fig:stab_points_4}\includegraphics[width=0.25\columnwidth]{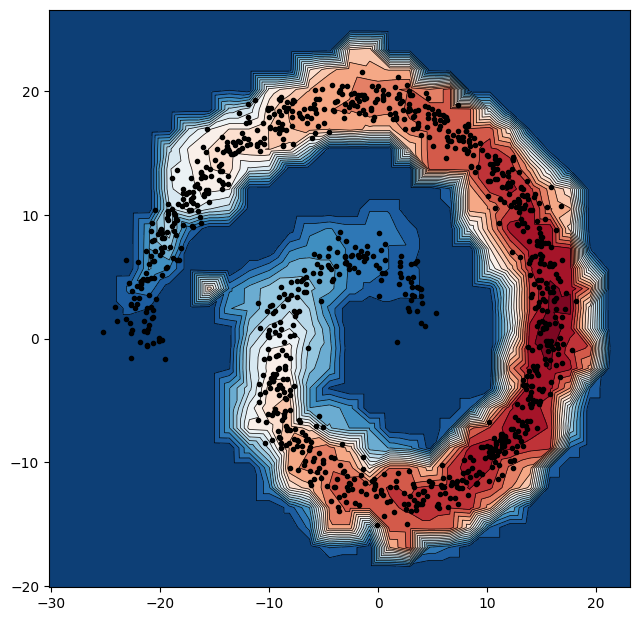}}
\caption{$LD$ using only $20\%$ of points (200 points) on simulated spiral dataset of 1000 points. The contours of $LD$ level changes slightly between different tries, but in general, the proposed method captures well the general shape of distribution.} 
\label{fig:stab_points}
\vspace{-1\baselineskip}
\end{figure}

\subsubsection{Stability with respect to the hyperparameter $\alpha$}

\begin{figure}
  \centering
  \subfloat[$\alpha=3$]{\label{fig:stab_alpha_3}\includegraphics[width=0.25\columnwidth]{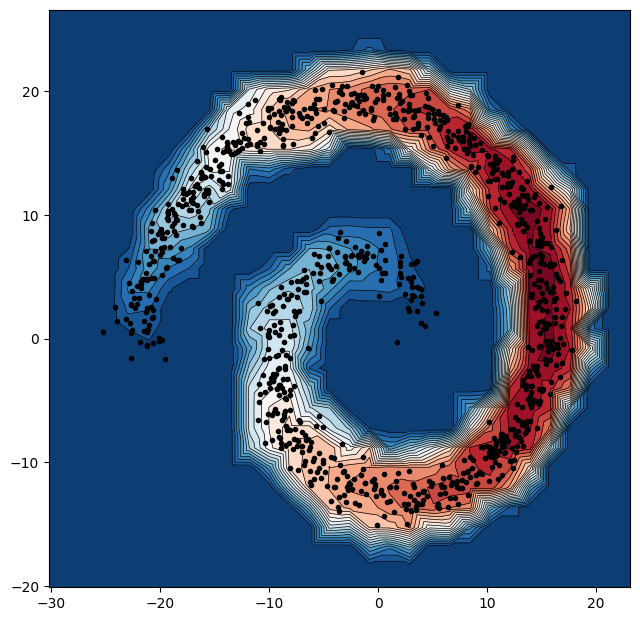}}
  \subfloat[$\alpha=5$]{\label{fig:stab_alpha_5}\includegraphics[width=0.25\columnwidth]{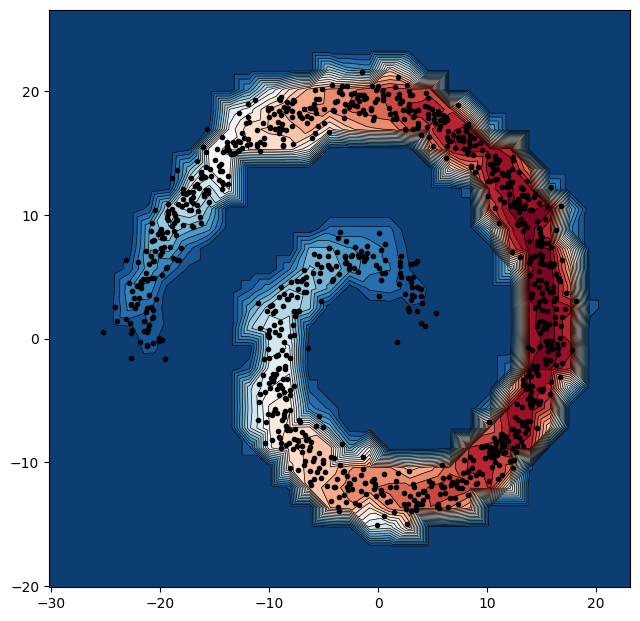}}
  \subfloat[$\alpha=10$]{\label{fig:stab_alpha_10}\includegraphics[width=0.25\columnwidth]{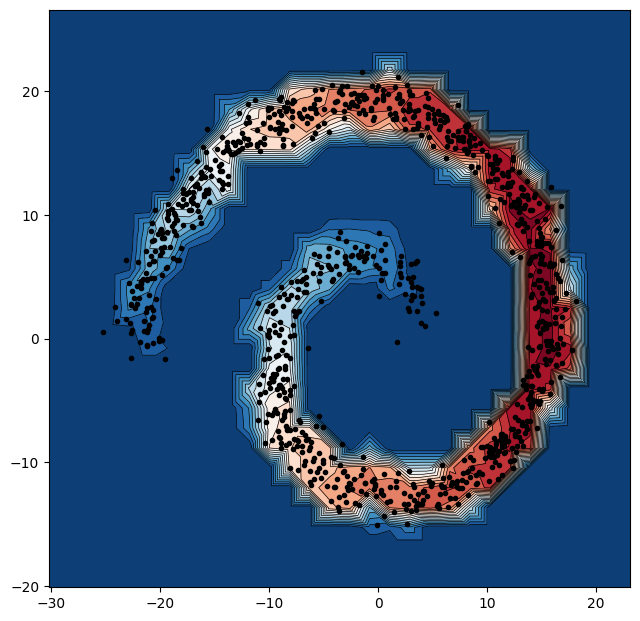}}
  \subfloat[$\alpha=15$]{\label{fig:stab_alpha_15}\includegraphics[width=0.25\columnwidth]{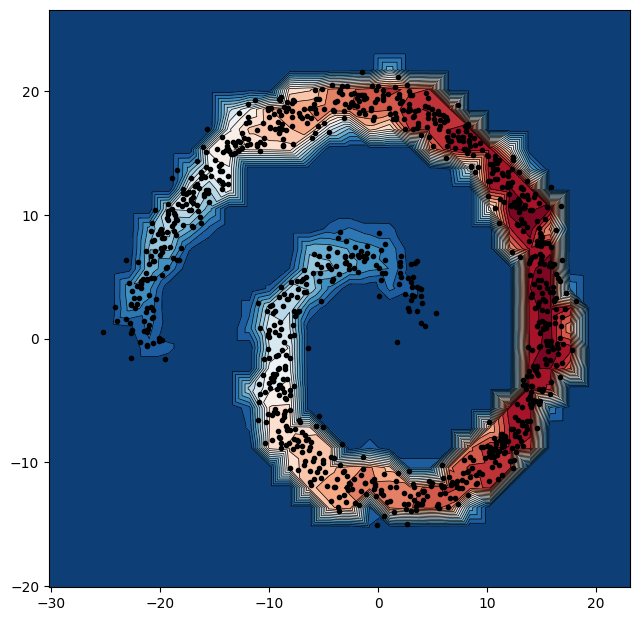}}
\caption{$LD$ applying different values of hyperparameter $\alpha$ in Sample Fermat Distance.  For different values of $\alpha$, the method always captures really well the distribution.} \label{fig:stab_alpha}
\vspace{-1.5\baselineskip}
\end{figure}

In  our method, only one hyperparameter  ($\alpha$), governing the Fermat distance needs to be chosen. It is therefore important to assess the stability of the method w.r.t. $\alpha$.   For this purpose, we experiment with different values of $\alpha>1$ (recall that $\alpha=1$ corresponds to the Euclidean distance). For each $\alpha \in \{3,5,10,15\}$, we test our method on the spiral dataset.  Results are shown in Fig. \ref{fig:stab_alpha}. The conclusion is that our method is very stable through different values of $\alpha$. Indeed, in the 4 cases, it always captures almost perfectly the dataset support,  which implies a strong stability of the method.
Of course this stability is only achieved in the proper range when $\alpha$ is large enough (See Fig. \ref{fig:min_path}).

\subsection{From LD to OOD uncertainty score} 
Our ultimate objective is to use $LD$ to provide an out-of-distribution uncertainty score. 
To do so, we apply $LD$ to the feature space $\mathcal{F}$ of our classification model.
Let $C$ be the number of separate clusters. Now, there are two ways for computing $LD$ of a new point: (1) All the clusters are considered as a sole distribution to compute $LD$; (2) Compute $LD$  w.r.t. the different clusters and then take the $max$ among the $LD$'s (i.e.  $LD$ w.r.t. the nearest cluster). It turns out that the first approach gives unsatisfying result as explained in \textbf{Appendix} \ref{one_or_multiple}. So, we adopt the second approach in this paper. More formally, let us denote $\widehat{LD}(\Phi (x), \mathcal{C}_i)$ the empirical LD of $x$ w.r.t. the $i^{th}$ cluster formed by training examples of class $i$ (in the feature space). Then, the confidence score of $x$ is defined as
\begin{equation}
    S(x) := \max_{i} \widehat{LD}(\Phi (x), \mathcal{C}_i) \ .
    \label{eq:define_score}
\end{equation}

\subsection{Reducing complexity for computing $LD$} 
\label{reduce}

From Eq. \eqref{eq:emp_ld}, we can deduce that the complexity of calculating $LD$ for a given point is more than $O(CN^2)$ ($C$ is the number of classes, $N$ is number of examples in each class).  It is therefore very interesting to reduce the number of inner points $N$ used to calculate $LD$ while maintaining good precision. Keeping only $n$ inner points among the $N$ initial ones, we then have 3 different straightforward strategies: 
\begin{itemize}
    \item \textbf{I. Random.} Randomly sample  without replacement $n$ points among $N$ intial points.
    \item  \textbf{II. K-mean/center.} We want the $n$ points to cover well the support of the initial sample. Hence, we first apply a \textit{K-mean clustering} with $n$ centroids on the $N$ points. Then, the $n$ resulted centroids are used as inner points. 
    \item \textbf{III.  K-mean/center+.} Same as strategy \textbf{II}, but instead of using directly the centroids, we use the inner point closest to each centroid.  
\end{itemize}

We test and discuss about these strategies in Appendix \ref{sec:reduce_strategy}. It turns out that \textit{K-mean / Center} outperforms the two other strategies with a very small number of inner points $n$. Indeed, \textit{K-mean / Center} has a regularization effect from averaging points (for calculating centroids). We conjecture that this effect makes the method much more stable, and also facilitates the capture of the overall shape of the cluster by avoiding certain irregular points that could have a negative impact on the estimate of $LD$. We refer to Appendix \ref{sec:reduce_strategy} for more discussion and detailed experiment. So, for the rest of this paper, we use strategy \textit{K-mean / Center}.

\section{Experiments on Neural Networks}
\label{exp}
We first evaluate our method on the two-moon dataset. Then, we evaluate on 2 benchmarks FashionMNIST/MNIST and CIFAR10/SVHN the ability of our method for the detection of Out-of-distribution. Besides, we also evaluate the consistency property of our uncertainty score as presented in introduction section (shown in Fig. \ref{fig:consistency_curve_fashion_mnist} and \ref{fig:consistency_curve_cifar10}).
Without further mention, we fix $\alpha=7$ for all experiments.

\textbf{Experiment Setup: }For a fair comparison, we use the same model architectures as in the previous work of \citet{van2020uncertainty} for the two experiments FashionMNIST/MNIST and CIFAR10/SVHN. More details about the  models and the training schemes can be found in the \textbf{Appendix} \ref{exp_details}. Moreover, as in \citet{van2020uncertainty}, at test time, we use the statistics (mean and standard deviation) of the training set (i.e. FashionMNIST or CIFAR10 in our case). Indeed, these statistics are used both in the \textit{Batch normalization} layers and in the  data normalization process  (both for  OOD and for ID set).

\subsection{How is the input distribution represented in the feature space of \textit{softmax} model?} \label{sec:two_moon}
We apply our method on the feature space $\mathcal{F}$ of \textit{softmax} model with the assumption that the input distribution is properly represented in $\mathcal{F}$. To assess this assumption, we first perform experiment on the two-moon dataset consisting of 2 classes, each having a  moon shape. We train a neural network with 2 hidden layers for classification\footnote{More details can be found in the \textbf{Appendix} \ref{exp_details}.}. After training, the model parameters are fixed and different methods for uncertainty evaluation are applied in $\mathcal{F}$.

\begin{figure}[ht]
  \centering
  \subfloat[\tiny{LD,$\alpha=3$}]{\label{fig:two_moon_LD_3}\includegraphics[width=0.25\columnwidth]{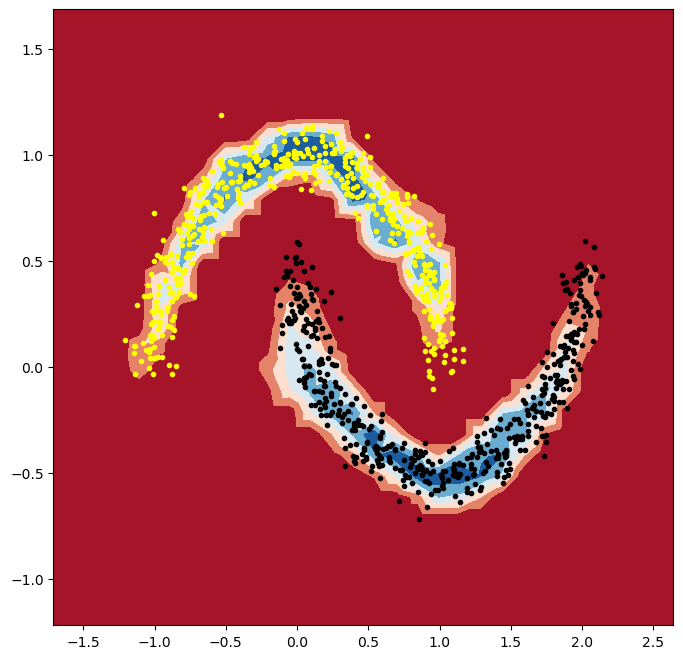}}
  \subfloat[\tiny{LD,$\alpha=10$}]{\label{fig:two_moon_LD_10}\includegraphics[width=0.25\columnwidth]{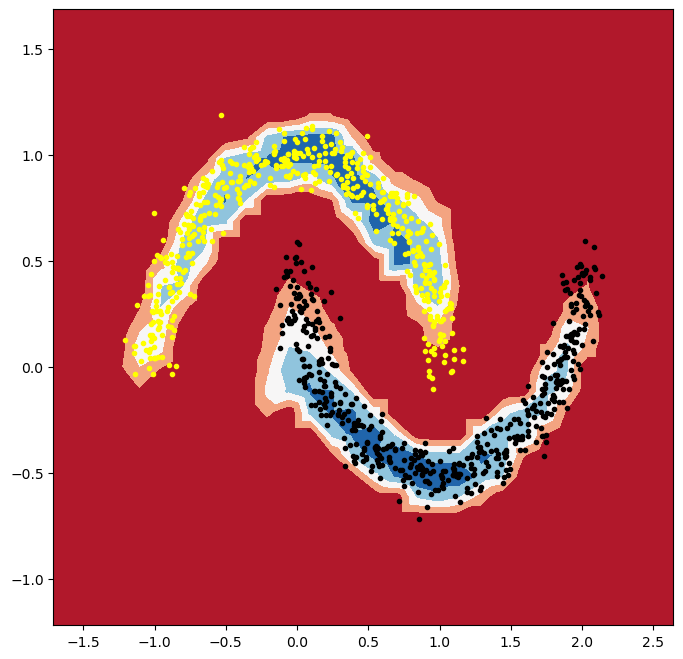}}
  \subfloat[\tiny{LD,$\alpha=15$}]{\label{fig:two_moon_LD_15}\includegraphics[width=0.25\columnwidth]{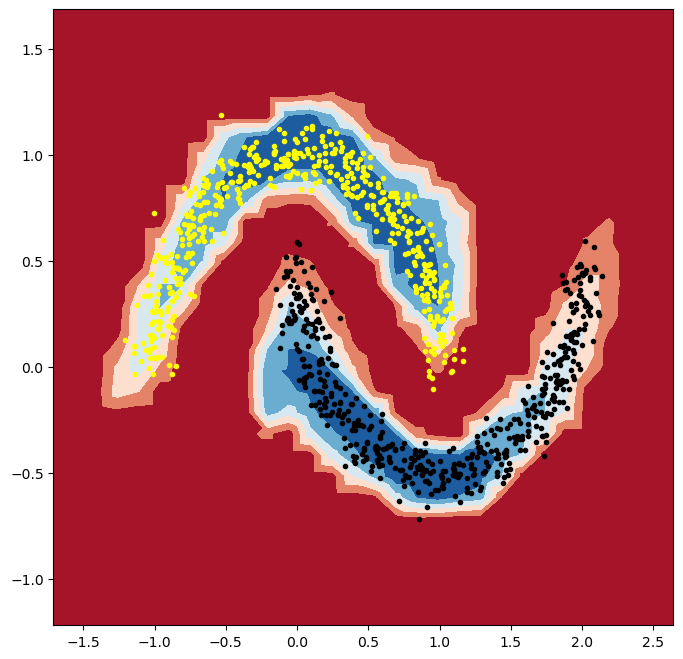}}\\
  \vspace{-1\baselineskip}
    \subfloat[\tiny{Euclidean}]{\label{fig:two_moon_euclidean}\includegraphics[width=0.25\columnwidth]{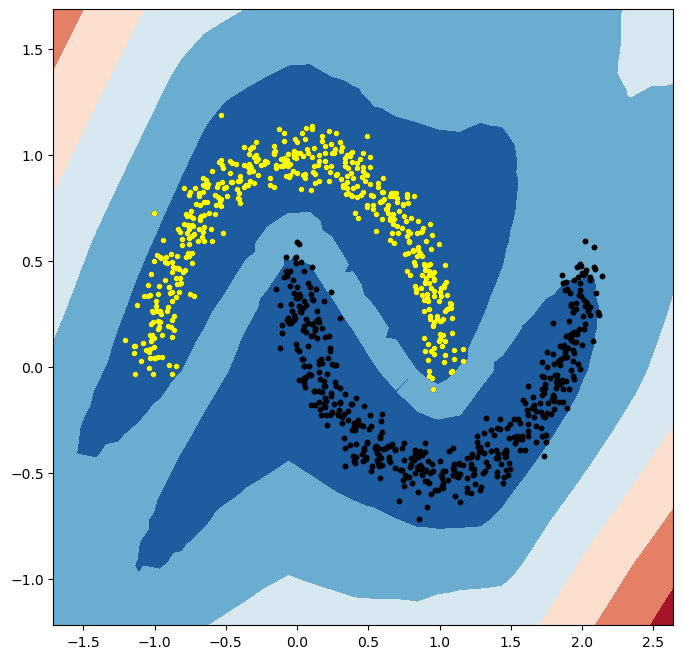}}
  \subfloat[\tiny{Mahalanobis}]{\label{fig:two_moon_gaussian}\includegraphics[width=0.25\columnwidth]{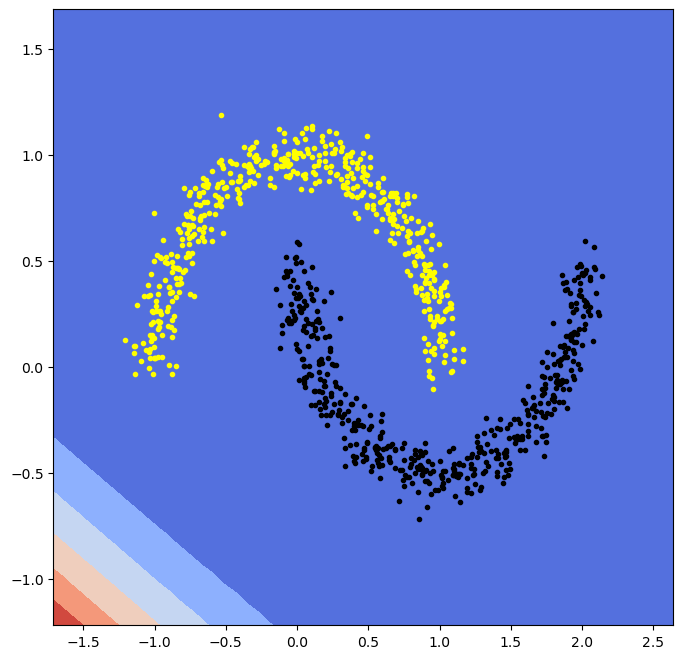}}
  \subfloat[\tiny{Entropy of predicted probability}]{\label{fig:two_moon_entropy}\includegraphics[width=0.25\columnwidth]{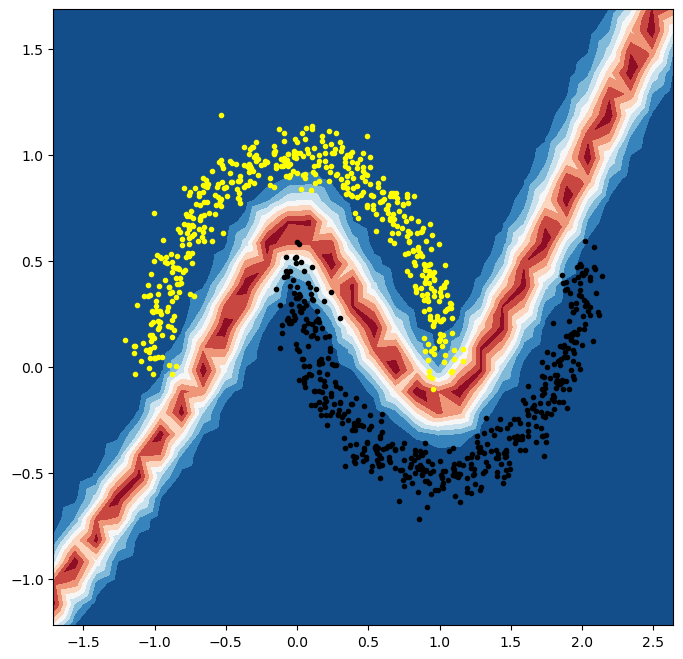}}
\caption{Methods for uncertainty estimation applied on the same neural net trained to classify 2 classes in moon-shape (represented by yellow and black points respectively). Uncertainty estimations are computed based solely on the feature space of the model without seeing directly the inputs. Our method (Fig. \ref{fig:two_moon_LD_3}, \ref{fig:two_moon_LD_10} and \ref{fig:two_moon_LD_15}) gives excellent results and much better than other methods.}\label{fig:two_moon_exp}
\vspace{-1\baselineskip}
\end{figure}

One popular way to provide an uncertainty score is to use the predictive distribution entropy\footnote{The entropy of a predicted probability $p \in \BBr^C$ is calculated as $H(p) = -\sum_{i=1}^C p_i\log(p_i)$, with $\sum_{i=1}^C p_i=1$ and $0\leq p_i \leq 1$.}. It is maximized when the distribution is uniform. In this example, predictive distribution entropy is high only in a boundary zone  (Fig. \ref{fig:two_moon_entropy}).  This is to be expected, as the model is trained to learn a boundary between the two classes. Nevertheless, it is  desired to assign a high uncertainty to the region without training data.  Indeed, it might be  too risky to make decision in these zones, especially in critical applications.

\textbf{Is Gaussian prior suitable?} We consider the methods of Euclidean distance (Fig. \ref{fig:two_moon_euclidean}) and Mahalanobis distance (Fig. \ref{fig:two_moon_gaussian}). For the Euclidean distance method, we compute the distance  to the centroids of the different clusters (in $\mathcal{F}$) and then we take its minimum. For the Mahalanobis distance method we proceed in the same way  but using the metric based on the covariance matrix. Surprisingly, in this example, the crude use of Euclidean distance seems to better capture the input distribution than the use of the  Mahalanobis distance (failing miserably on this dataset). This suggests that the distribution of clusters in feature space is more complicated than the Gaussian one.

This remark shows the necessity to have a method able to capture better the distribution. $LD$   can capture impressively well the zone where we have training data (Fig. \ref{fig:two_moon_LD_3}, \ref{fig:two_moon_LD_10} and \ref{fig:two_moon_LD_15} corresponding to $\alpha =3,\,10$ and $15$). Hence, $LD$ is able to pin down clusters with a complex support shape in feature space. Furthermore, we intentionally use 3 values for $\alpha$ with large gaps to show the stability w.r.t. $\alpha$. 


\subsection{FashionMNIST vs MNIST} \label{sec:fashion_mnist}

We perform five different runs to train classification models on the data set  FashionMNIST \cite{xiao2017fashionmnist}. 

Firstly, we evaluate our method by studying the separation capacity between the test set of FashionMNIST and of MNIST \cite{lecun1998mnist} based on \textit{AUROC} score.
Results are reported in Table \ref{table:auroc_mnist}. We first compare our method to Euclidean and Mahalanobis distance method \cite{lee2018simple} (that are explained in Section \ref{sec:two_moon}). Notice that our method outperforms these two distance-based methods. Notice also that Euclidean distance gives slightly better AUROC score than Mahalanobis one. This might result from the fact that the distributions are far to be Gaussian, confirming once again that Gaussian assumption does not hold. A more sophisticated method called DUQ stands on a devoted neural architecture (RBF network) (see \cite{van2020uncertainty}).
This particular type of model is much more difficult to train and so generally does not preserve the accuracy of the main classification task (compared to standard \textit{softmax} models).
Once again, our method outperforms this method. This indicates that our method  measures a natural ``depth'' directly in the feature space without the need of changing  completely the model as in DUQ method. Another popular method is Deep Ensembles in which one trains and applies many independently-trained models for the same task. Consequently, this approach needs more resource both at training and inference times.
Despite its heavy demanding of resource, our method outperforms this approach in this experiment. A more advanced method for density estimation is LL ratio \cite{ren2019likelihood}. In this method, one needs to train two supplementary generative models to estimate distributions.  A first model is trained on ID data and a second one is trained on perturbed inputs.
This method needs an adequate noise in the way that the perturbed inputs contain only background information. Moreover, one needs to really carefully train these 2 generative models so that they can reflect the true underlying input density. With this complex process, this method gives better AUROC score than ours (in this experiment). However, we will show that our method outperforms it in the next experiment.

\begin{table}[]
\caption{Results on FashionMNIST, with MNIST as OOD set. Results marked by ($\Box$) are extracted from \citet{ren2019likelihood} and ($\triangle$) are extracted from  \citet{van2020uncertainty}. Deep Ensembles by \citet{lakshminarayanan2017simple}, Mahalanobis Distance by \citet{lee2018simple}, LL ratio by \citet{ren2019likelihood}, DUQ by  \citet{van2020uncertainty}.}
\label{table:auroc_mnist}
\begin{tabular}{|l|l|l|}
\hline
Charateristics                                                                & Method                                                              & AUROC \\ \hline
\multirow{3}{*}{\begin{tabular}[c]{@{}l@{}}No impact\\ on original model\end{tabular}} & LD (our method)       & $0.971 \scriptstyle \pm 0.001$     \\ \cline{2-3} 
& \begin{tabular}[c]{@{}l@{}}Euclidean\\ Distance\end{tabular}        & $0.943 \scriptstyle \pm 0.009$    \\ \cline{2-3} 
& \begin{tabular}[c]{@{}l@{}}Mahalanobis\\ Distance ($\Box$)\end{tabular}      & 0.942    \\ \hline
\begin{tabular}[c]{@{}l@{}}Use particular type\\ of model difficult\\ to train\end{tabular}                  & DUQ ($\triangle$)                                                                 & 0.955    \\ \hline
\begin{tabular}[c]{@{}l@{}}Need to train\\ many models\end{tabular}                         & \begin{tabular}[c]{@{}l@{}}Deep Ensembles\\ (5 Models) ($\triangle$) \end{tabular} & 0.861    \\ \hline
\begin{tabular}[c]{@{}l@{}}Need to train extra\\ generative models \end{tabular}     & LL ratio ($\triangle$)                                                           & 0.994    \\ \hline
\end{tabular}
\vskip -0.1in
\end{table}

\begin{figure}[ht]
\centering
    \includegraphics[width=0.6\columnwidth]{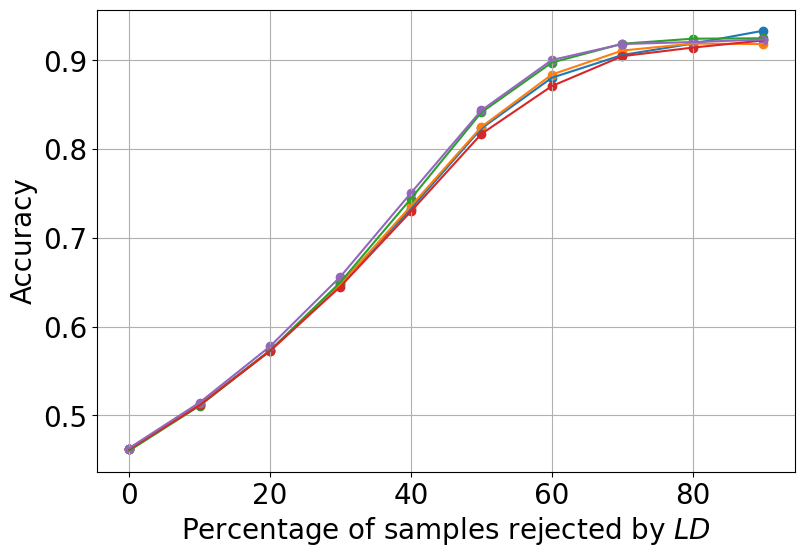}
\caption{Consistency curve FashionMNIST/MNIST for 5 runs (each curve corresponds to an independently trained model).} \label{fig:consistency_curve_fashion_mnist}
\end{figure}

{\bf Consistency curve.} Next, we evaluate the usefulness of our uncertainty estimation for decision making. Following some previous works (e.g. \citet{lakshminarayanan2017simple}, \citet{van2020uncertainty}), we compound test set of FashionMNIST and MNIST together and all the data from MNIST are considered to be incorrectly predicted by the model. Then, a certain percentage of data is rejected based on their $LD$. If $LD$ is an appropriate indicator for prediction uncertainty, then accuracy on the retained data has to be an increasing function of the rejection percentage. We call the resulted curve \textit{consistency curve}. The results for five runs are depicted in Fig. \ref{fig:consistency_curve_fashion_mnist}. We see that the curves are always increasing over 5 runs. Hence, $LD$ is a good measure for uncertainty estimation. Moreover, the 5 curves are very similar. This implies the stability of our method.

\subsection{CIFAR10 vs SVHN} \label{sec:svhn_cifar10}

We perform the same study as in the previous experiment. Nevertheless,  this time we train the model on the training set CIFAR10 \cite{cifar10} and  then we test on the test sets of CIFAR10 and of SVHN \cite{netzer2011reading}. AUROC scores are reported in Table \ref{table:auroc_cifar10}. We see that our method outperforms the other baseline methods. Especially, this time our method outperforms Deep Ensembles with 3 and 5 models that are costly to train. In addition, our method also outperforms the LL ratio method, although this last method requires the careful training of two additional generative models only for density estimation, which is a difficult task in itself. This reinforces our intuition that we can work directly in the feature space of the base model (i.e. the model for the main classification problem), rather than to fit supplementary models just for density estimation.  

\begin{table}[]
\caption{Results on CIFAR10, with SVHN as OOD set. Our method outperforms other baseline methods. Results marked by ($\Box$) are extracted from  \citet{van2020uncertainty} and ($\triangle$) extracted from \citet{sun2022out}. Deep Ensembles by \citet{lakshminarayanan2017simple}, LL ratio by \citet{ren2019likelihood}, DUQ by  \citet{van2020uncertainty}, Energy by \citet{liu2020energy}, KNN by \citet{sun2022out}.}
\label{table:auroc_cifar10}
\begin{tabular}{|l|l|l|l|}
\hline
\begin{tabular}[c]{@{}l@{}} Extra \\ data\footnote{Yes means there is the need of auxiliary OOD data.}
\end{tabular}
& Intrusive\footnote{The term \textit{intrusive} means that in order to get the result, one needs to change/fine-tune the original model (such as DUQ and Energy method), or to train supplementary models (such as Deep Ensembles or LL ratio method).}
                       & Method         & AUROC \\ \hline
& \textbf{No}          & LD (our method)      & {$\mathbf{0.936} \scriptstyle \pm 0.006$} \\ \cline{2-4} 
& Yes & DUQ ($\Box$)        & $0.927 \scriptstyle \pm 0.013$ \\ \cline{2-4} 
& Yes & \begin{tabular}[c]{@{}l@{}}Deep Ensembles\\ (3 models) ($\Box$)\end{tabular} & $0.926 \scriptstyle \pm 0.010 $ \\ \cline{2-4} 
\multirow{-4}{*}{\begin{tabular}[c]{@{}l@{}} {\bf No}
\end{tabular}}
& Yes & \begin{tabular}[c]{@{}l@{}}Deep Ensemble\\ (5 models) ($\Box$) \end{tabular}  & $0.933 \scriptstyle \pm 0.008$ \\ \hline
& Yes & LL ratio ($\Box$)         & 0.930 \\ \cline{2-4} 
& Yes & Energy($\triangle$)       & 0.912 \\ \cline{2-4} 
\multirow{-3}{*}{\begin{tabular}[c]{@{}l@{}}
Yes
\end{tabular}}
& \textbf{No}  & KNN ({\footnotesize ours})    & $0.926 \scriptstyle \pm 0.002 $ \\ \hline
\end{tabular}
\vspace{-0.5\baselineskip}
\end{table}

In the Energy method \cite{liu2020energy}, the authors show that \textit{softmax} model is a special case of Energy-based model \cite{lecun2006tutorial}. Based on this insight, one can derive energy from logits of \textit{softmax} layer and then use the energy  as a criteria to distinguish ID from OOD data.
However,  one needs to carefully tune the model using supplementary OOD data to increase the margin of energy between ID and OOD data. In this experiment, our method outperforms this method. This implies that capturing distribution in the feature space using our method is more effective than working on the logits of \textit{softmax} layer despite its interpretation form energy-based point of view.

More recently, deep nearest neighbors method was proposed by \citet{sun2022out} (denoted as \textit{KNN} in Table \ref{table:auroc_cifar10}). In this method, one first computes the (Euclidean) distance of the considered point to all the training points in the feature space (after normalizing the features using $l^2$-norm). Then, the distance to the $k^{th}$ nearest neighbor is the criteria to distinguish ID from OOD data. $k$ is a hyperparameter chosen based on a validation process using an auxiliary OOD data. The AUROC score in the original paper is $0.959$. However, when we apply this method on our trained models (over 5 runs), our method outperforms this method (Table \ref{table:auroc_cifar10}). As the 2 methods are applied in the same feature space, it would be fair to say that our method can capture better the distribution than \textit{KNN} method.  This is not surprising, since this method is rather heuristic and not based on clear statistical or probabilistic foundations, whereas ours  is based on well-developed statistical notion which is \textit{LD}.

\begin{figure}[ht]
\centering
    \includegraphics[width=0.6\columnwidth]{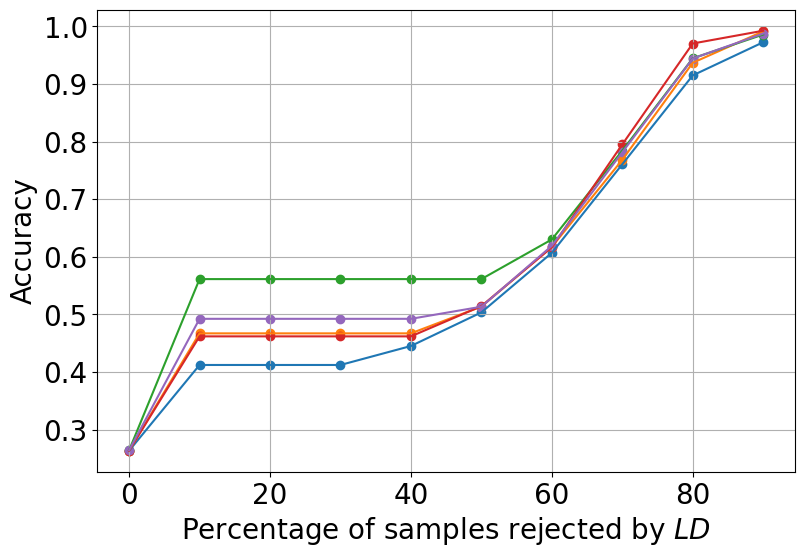}
\caption{Consistency curve CIFAR10/SVHN for 5 runs (each curve corresponds to an independently trained model).} \label{fig:consistency_curve_cifar10}
\vspace{-1\baselineskip}
\end{figure}

Next, we plot the consistency curve for the 5 runs with the combined test set of CIFAR10 and of SVHN, where the data in SVHN test set are considered incorrect. (Fig. \ref{fig:consistency_curve_cifar10}). Once again, accuracy is always an increasing function of the rejected percentage of the data based on $LD$. This confirms again that $LD$ is an appropriate indicator for uncertainty estimation and so it is useful for decision making.

\section{Discussion}
\label{discussion}

In this work, we aim to show that with a sharp enough method for capturing the distribution in the feature space, we should have access to an appropriate indicator for uncertainty estimation. In the two-moon example where we train a model for classification, Euclidean and Mahalanobis distances (i.e. Gaussian fitting) fail to capture the input distribution (Fig. \ref{fig:two_moon_euclidean} and \ref{fig:two_moon_gaussian}). So, one could have conjectured that the feature space itself cannot conserve the input distribution. However, when applying our method, the input distribution is totally captured and perfectly represented (Fig. \ref{fig:two_moon_LD_3} and \ref{fig:two_moon_LD_10}). This indicates that the problem does not reside in the properness of the feature space itself but in the methods capturing the distribution. Through more challenging examples on FashionMNIST/MNIST and CIFAR10/SVHN, we once again confirm this last statement as our method gives AUROC scores that are competitive or better than those of other methods. 

In some prior works, one makes ID and OOD data more separable in the feature space by applying techniques such as: performing some preprocessing techniques on the inputs (e.g. \citet{liang2017enhancing}, \citet{lakshminarayanan2017simple}); fine-tuning neural net using OOD data (e.g. \citet{liu2020energy}); applying some particular loss function on the feature space (e.g. \citet{sun2022out}). These techniques are orthogonal to our method and so they can be combined with ours. Hence, these techniques are not considered here. Indeed,  we aim to highlight the importance of the method capturing the distribution.
Combining these techniques with our method is out-of-scope of this paper and will be studied in a future work. 

\section{Conclusion}
\label{conclusion}
In this work, we use the statistical notion of Lens Depth combined with a modified version of the sample Fermat distance. This combination captures naturally the shape and density of the input distribution. This is not the case with many previously proposed methods, which assume a prior distribution or use additional models with trainable parameters, or even modify the mechanism of the training process.
Our method is non-parametric and non-intrusive. Through a toy dataset as well as experiments conducted on Deep Neural Networks, we show that our method  adapts very well to many cases. Hence, our work is opening new tracks for non-parametric methods capturing the input distribution  to quantify uncertainty in the context of Deep Learning. For future work, it would be interesting both to have an efficient algorithm for computing Lens Depth with some error bound and to investigate more the impact of the hyper-parameter $\alpha$. 

Finally notice that, while we were focused on neural networks, any classification model with a feature space, e.g. kernel methods, can benefit from our framework.






\bibliography{paper}
\bibliographystyle{icml2024}



\newpage
\ 
\newpage 

\appendix

\section{Experimental Details}
\label{exp_details}
All experiments related to neural networks are implemented in Pytorch \textit{2.0.1+cuda}, with its default initialization.
\subsection{Two-moons} \label{seq:append_two_moon}
For generating two-moon dataset, we use package scikit-learn \cite{scikit-learn}, with noise 0.07, random state 1, and 1000 samples.

For model, we construct a simple fully connected network with 2 hidden layers, each gives output of dimension 20. First hidden layer is followed by non-linearity \textit{ReLU}. We train model for 160 epochs using Adam optimizer, learning rate $10^{-3}$, other parameters are set by default of Pytorch package.

\subsection{FashionMNIST} \label{seq:append_mnist}
For a fair comparison, we follow exactly the same CNN architecture proposed in \cite{van2020uncertainty} and the same training scheme with only one minor modification: the dimension of penultimate layer is 25 instead of 256 for efficient computation related to \textit{LD}. We observe this modification has no impact on accuracy of model. We refer reader to \cite{van2020uncertainty} for details. From training set, we randomly split 80:20 to have training data and validatation. We choose the best model based on accuracy on validation set. Test accuracy after training over 5 runs is $92.35 \% \pm 0.19$.

For estimating $\widehat{LD}$, we use 1500 training examples for each class based on results of the experiment in Section \ref{sec:reduce_strategy}. We observed that applying the method on normalized feature vectors (L2-norms) (which is the reported result) gives slightly better result than applying directly on the feature vectors. The method is applied on the test set of FashionMNIST consisting of 10,000 images and the test set of MNIST also consisting of 10,000 images).

\subsection{CIFAR10} \label{seq:append_cifar10}
 We use ResNet-18 model implemented by \cite{devries2017cutout} with a minor modification and training scheme of the same authors. More specifically, after Global Average Pooling layer of CNN, we add a layer of output dimension of 25 instead of 256 proposed by \cite{van2020uncertainty} before softmax layer. For training model, we use SGD optimizer with nesterov momentum, learning rate 0.1 (decayed by factor of 5 at epochs 60,120 and 160), momentum 0.9, weight decay $5 \cdot 10^{-4}$.Model is trained for 200 epochs. We train model on the full training set (i.e. no validation set) and save the last model, i.e. the model at epoch 200. Test accuracy after training over 5 runs is $0.950 \pm 0.001$.

As there are many more images to test compared to the previous experiment, we use only 1000 training images in each class for estimate $LD$ using \textit{K-mean/Center} strategy to have a reasonable run time. We tested the method on normalized and non-normalized feature vectors and observed no significant difference. The reported result is on non-normalized vectors. Notice that the method is applied on test sets of CIFAR10 consisting of 10,000 images and of SVHN consisting of 26,032 images.

\section{More qualitative results of Fermat Lens Depth on spiral dataset using $20\%$ points} \label{sec:append_stable_points}
\label{stab_points}
\begin{figure}[h!]
  \centering
  \subfloat[$1^{st}$ try]{\label{fig:stab_points_app_1}\includegraphics[width=0.25\columnwidth]{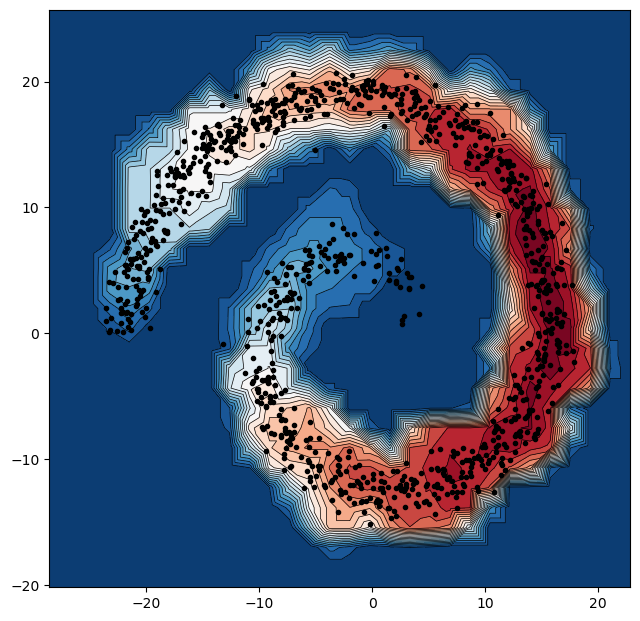}}
  \subfloat[$2^{nd}$ try]{\label{fig:stab_points_app_2}\includegraphics[width=0.25\columnwidth]{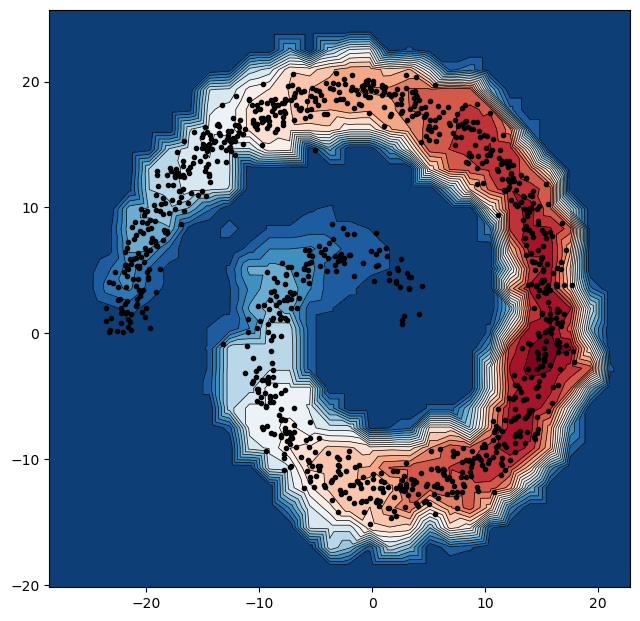}}
  \subfloat[$3^{rd}$ try]{\label{fig:stab_points_app_3}\includegraphics[width=0.25\columnwidth]{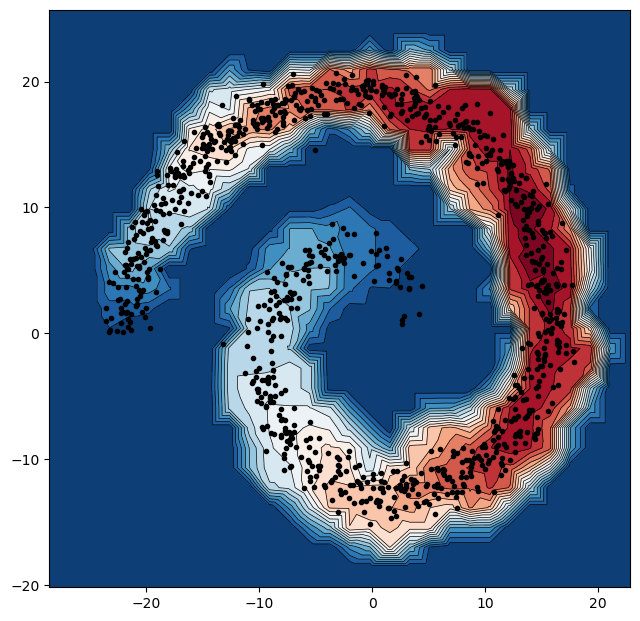}}
  \subfloat[$4^{th}$ try]{\label{fig:stab_points_app_4}\includegraphics[width=0.25\columnwidth]{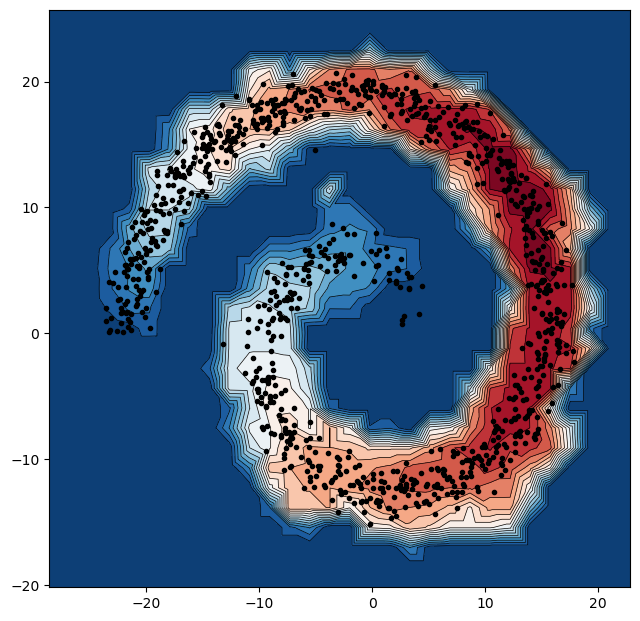}}\\
  \vspace{-1\baselineskip}
  \subfloat[$5^{th}$ try]{\label{fig:stab_points_app_5}\includegraphics[width=0.25\columnwidth]{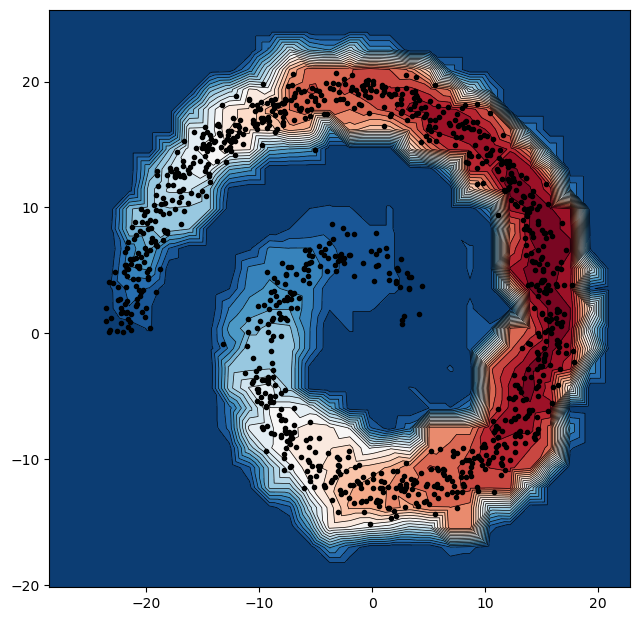}}
  \subfloat[$6^{th}$ try]{\label{fig:stab_points_app_6}\includegraphics[width=0.25\columnwidth]{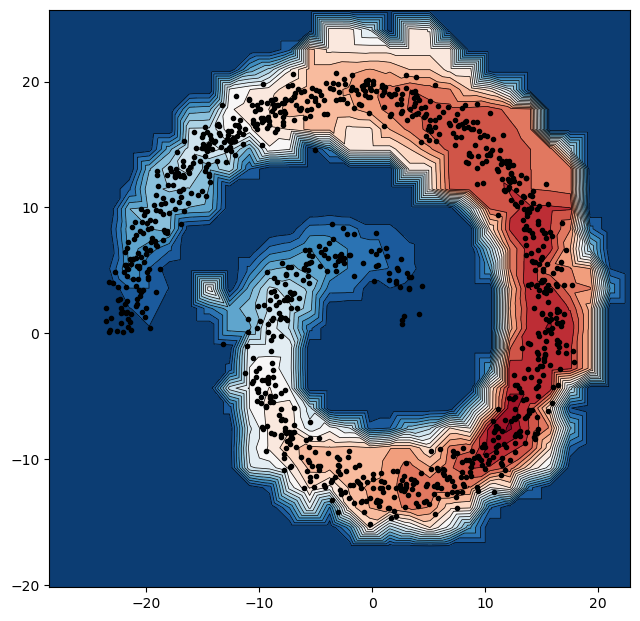}}
  \subfloat[$7^{th}$ try]{\label{fig:stab_points_app_7}\includegraphics[width=0.25\columnwidth]{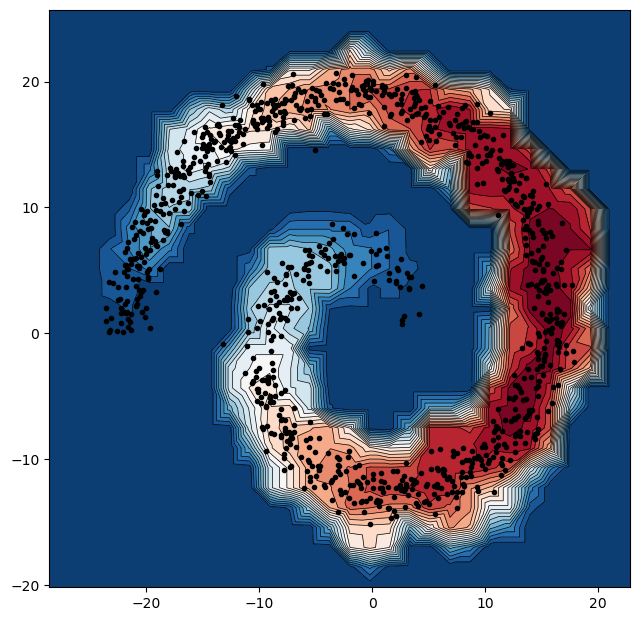}}
  \subfloat[$8^{th}$ try]{\label{fig:stab_points_app_8}\includegraphics[width=0.25\columnwidth]{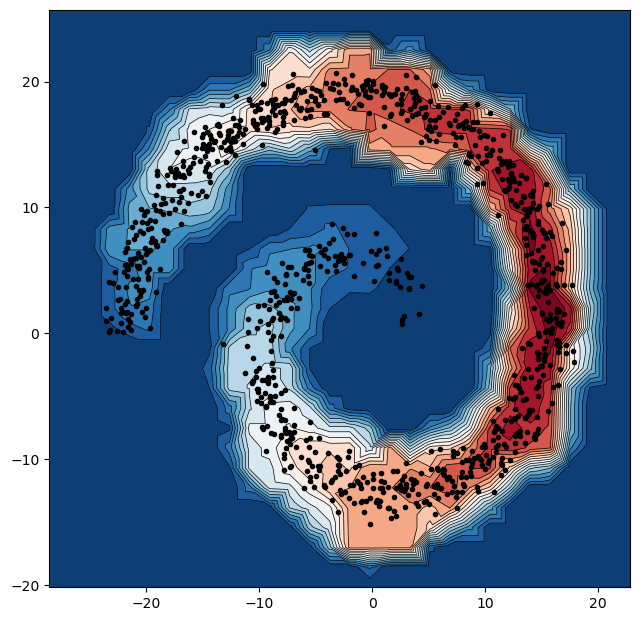}}\\
  \vspace{-1\baselineskip}
  \subfloat[$9^{th}$ try]{\label{fig:stab_points_app_9}\includegraphics[width=0.25\columnwidth]{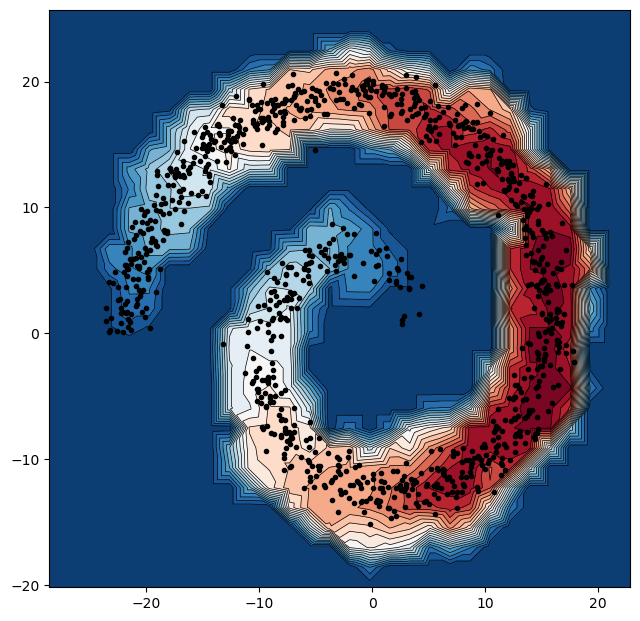}}
  \subfloat[$10^{th}$ try]{\label{fig:stab_points_app_10}\includegraphics[width=0.25\columnwidth]{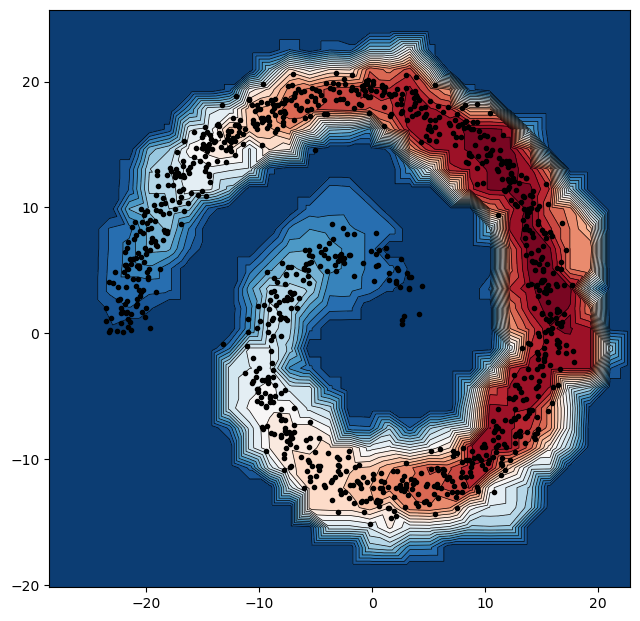}}\\
  \vspace{-1\baselineskip}
\caption{$LD$ using only $20\%$ of points (200 points) on simulated spiral dataset of 1000 points over 10 runs. We see that the contours of $LD$ level changes slightly between different tries, but in general, the proposed method captures well the general form of distribution. Note that the points presented in the plot are the full dataset of 1000 points.} \label{fig:stab_points_app}
\end{figure}

We use $LD$ using only $20\%$ of points (200 points) on simulated spiral dataset of 1000 points over 10 runs for qualitatively evaluating stable of the method w.r.t. number of points.Results in Fig.\ref{fig:stab_points_app}.

\section{Proof of Proposition 1}
\label{proof}
\begin{proof}
By definition, $q(q(x))=q(x)$ and $q(y)=y$ as $q(x), y \in Q$, so the closest point to them in $Q$ is themselves. Hence, according to Eq.\ref{eq:sample_fermat}, it is obvious that $D_{Q,\alpha}(x,y) = D_{Q,\alpha}(q(x),y), \forall x \in \mathbb{R}^d \quad \text{and} \quad y \in Q$. Applying this sample Fermat distance to Eq.\ref{eq:emp_ld} to compute empirical $LD$,  we obtain $\widehat{LD}(x)=\widehat{LD}(q(x))$.
\end{proof}

\section{One or multiple clusters?}
\label{one_or_multiple}
Our ultimate objective is to use $LD$ for finding out-of-distribution data associated with an uncertainty (or equivalently confidence) score. For this purpose, we apply $LD$ in feature space of \textit{softmax} model. More concretely, we apply in the activation of penultimate layer right before \textit{softmax} layer. In this setup where we have different separate clusters, one important question is: \textit{Can we simply consider them as one distribution for computing LD?}

\begin{figure}[ht]
  \centering
  \subfloat[Consider 3 clusters as one distribution]{\label{fig:ld_multiple_gaussians}\includegraphics[width=0.4\columnwidth]{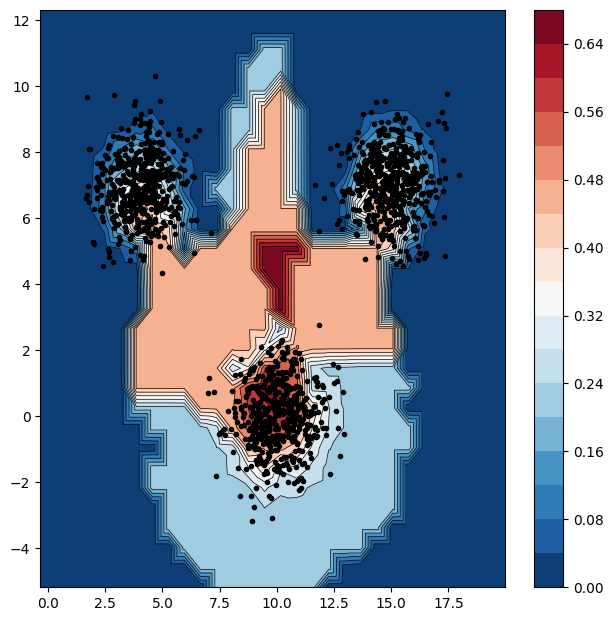}}
  \subfloat[\textit{Max} of $LD$ w.r.t to different clusters]{\label{fig:ltiple_gaussians_separate}\includegraphics[width=0.4\columnwidth]{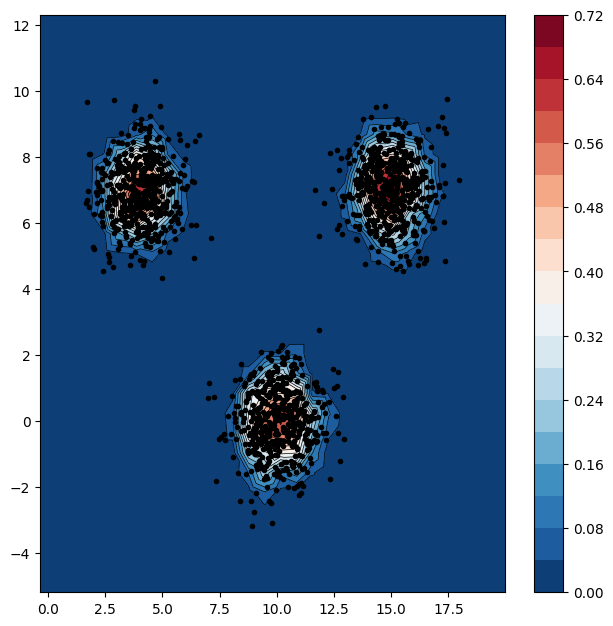}}
\caption{$LD$ computed by 2 ways: (a)Consider 3 clusters as one distribution w.r.t which one compute $LD$ and (b) Compute $LD$ w.r.t 3 separate clusters than final $LD$ is computed as the \textit{max} among the 3 $LD$'s computed} \label{fig:ld_multiple_clusters}
\vspace{-1\baselineskip}
\end{figure}

To answer this question, we simulate a dataset composed of 3 separate Gaussian clusters and then we compute $LD$ w.r.t this dataset by consider them as one distribution. The result is shown in Fig.\ref{fig:ld_multiple_gaussians}. We see that the result is not good: value of $LD$ is large for zones lying between clusters whereas in this case, we want $LD$ to be large only in 3 cluster and NOT in the zones lying between them. So, the solution for this problem is quite straightforward: we compute $LD$ of a points w.r.t different clusters, than the final $LD$ of that point is considered as  its $LD$ w.r.t to the closest cluster, i.e. we take the \textit{max} among computed $LD$'s. The result is shown in Fig.\ref{fig:ltiple_gaussians_separate}. We see that the result now is much bette: $LD$ is only large in the zones of dataset which are 3 clusters.

\section{Effectiveness of Reduced Lens Depth} 
\label{sec:reduce_strategy}

In this section, we evaluate the effectiveness of the 3 strategies discussed in Section \ref{reduce} to reduce the computing complexity of $LD$. We evaluate the quality of each strategy by measuring how well the ID can be separated from OOD set in term of the \textit{AUROC} metric\footnote{AUROC is equal to 1 when the data are perfectly separable.}. The pair FashionMNIST/MNIST is used to assess the suitability of the three strategies. This pair is much more difficult to handle than MNIST/NotMNIST as argued in previous works (e.g. \citet{van2020uncertainty}). For each strategy, we use $n \in \{500, 1000, 1500\}$ points for each class (each class contains 6000 training examples). Results are reported in Table \ref{table:compare_reduce}. In all cases, strategy \textbf{II} always gives the best result. Remarkably, with only 500 points, \textit{K-mean / Center} is already better than the two other strategies with 1500 points. 
\textit{K-mean / Center} has a regularization effect from averaging points (for calculating centroids). We conjecture that this effect makes the method much more stable, and also facilitates the capture of the overall shape of the cluster by avoiding certain irregular points that could have a negative impact on the estimate of $LD$.

Moreover, as number of points are increased from 500 to 1500, we observe no significant improvement in \textit{AUROC} score. This reinforces our conjecture about the impact of irregular points on estimation of $LD$ and furthermore, this remark implies that if the $n$ chosen points cover well enough the initial space occupied by the $N$ original points, then we only need to choose a very small percentage of points for a good estimation of $LD$. So, for the rest of this paper, we use strategy \textit{K-mean / Center}.
\begin{table}[t]
\caption{Comparing AUROC performance of strategies for reducing complexity of LD on FashionMNIST/MNIST}
\label{table:compare_reduce}
\vskip -0.5in
\begin{center}
\begin{small}
\begin{sc}
\begin{tabular}{lcccr}
\toprule
No. of training examples & 500 & 1000 & 1500 \\
\midrule
I. Random    & 0.9368 & 0.9426 &  0.9436 \\
II. K-mean / Center & \textbf{0.9543} & \textbf{0.9548} &  \textbf{0.9553} \\
III. K-mean / Center+  & 0.9475 & 0.9536 &  0.9537 \\
\bottomrule
\end{tabular}
\end{sc}
\end{small}
\end{center}
\vskip -0.1in
\vspace{-1\baselineskip}
\end{table}

\section{Voronoï cells}
\label{voronoi}
Suppose we have a finite number of distinct points in the plane,  referred as \textit{sites}, \textit{seeds} or \textit{generators}. Each seed has a corresponding region, called a Voronoï cell, made up of all the points in the plane closer to that seed than to any other. We refer to \cite{aurenhammer1991voronoi} for more details.

\end{document}